\documentclass{article} %
\usepackage{iclr2026_conference,times}

\usepackage{amsmath}

\usepackage{amsmath,amsfonts,bm}

\def\eqref#1{equation~\ref{#1}}

\def\1{\bm{1}}

\DeclareMathAlphabet{\mathsfit}{\encodingdefault}{\sfdefault}{m}{sl}
\SetMathAlphabet{\mathsfit}{bold}{\encodingdefault}{\sfdefault}{bx}{n}

\usepackage{hyperref}
\usepackage{url}
\usepackage{amssymb}            %
\usepackage{mathtools}          %
\usepackage{mathrsfs}           %
\usepackage[group-separator={,}]{siunitx}
\usepackage{graphicx}           %
\usepackage{caption, subcaption}         %
\usepackage[space]{grffile}     %
\usepackage{url}                %
\usepackage{enumitem}
\usepackage{float}

\usepackage[many]{tcolorbox}

\usepackage{lipsum}
\usepackage{wrapfig}
\usepackage{pbox}
\usepackage{array}
\usepackage{listings}
\usepackage{placeins}

\usepackage{amsthm}

\usepackage{uoftcolors}

\usepackage{uoftcolors}
\hypersetup{
  colorlinks=true,
  allcolors=uoftblue
}

\usepackage{todonotes}
\usepackage[normalem]{ulem}
\usepackage{cancel}
\usepackage{algorithm2e}
\makeatletter
\renewcommand{\@algocf@capt@plain}{above}%
\makeatother

\usepackage[capitalize,nameinlink,noabbrev]{cleveref} 
\crefformat{equation}{(#2#1#3)}

\usepackage{thm-restate}

\usepackage{uoftcolors}

\AtBeginDocument{}%
\AtBeginDocument{}%
\AtBeginDocument{}%
\AtBeginDocument{}%

\newcommand{\blue}[1]{{\color{uoftblue} #1}}
\newcommand{\red}[1]{{\color{uoftred} #1}}

\usepackage{xspace}

\renewcommand{\paragraph}[1]{\textbf{#1}~~~}

\definecolor{sb1}{HTML}{1f77b4}
\definecolor{sb2}{HTML}{ff7f0e}
\definecolor{sb3}{HTML}{2ca02c}
\definecolor{sb4}{HTML}{d62728}
\definecolor{sb5}{HTML}{9467bd}
\definecolor{sb6}{HTML}{8c564b}
\definecolor{sb7}{HTML}{e377c2}
\definecolor{sb8}{HTML}{7f7f7f}
\definecolor{sb9}{HTML}{bcbd22}
\definecolor{sb0}{HTML}{17becf}

\newcommand{\algoname}[0]{REPPO\xspace}
\newcommand{\fullalgoname}[0]{Relative Entropy Pathwise Policy Optimization
}

\title{Relative Entropy Pathwise \\Policy Optimization}

\author{Claas A Voelcker$^{1,2,*}$, Axel Brunnbauer$^{3,*}$, Marcel Hussing$^4$, Michal Nauman$^{5,6}$, \\
\textbf{Pieter Abbeel$^6$, Eric Eaton$^4$, Radu Grosu$^3$, Amir-massoud Farahmand$^{7,8,1}$,} \\
\textbf{Igor Gilitschenski$^{1,2}$} \\ 
$^1$University of Toronto, $^2$Vector Institute, $^3$TU Wien, $^4$University of Pennsylvania \\
$^5$University of Warsaw,
$^6$UC Berkeley, $^7$Polytechnique Montr\'eal, $^8$Mila -- Quebec AI Institute \\
$^*$ Authors contributed equally, correspondence to \url{cvoelcker@cs.toronto.edu},\\
\url{axel.brunnbauer@tuwien.ac.at}
}

\iclrfinalcopy %
\begin{document}

\maketitle

\begin{abstract}
Score-function based methods for policy learning, such as REINFORCE and PPO, have delivered strong results in game-playing and robotics, yet their high variance often undermines training stability.
Improving a policy through state-action value functions, for example by differentiating Q with regard to the policy, alleviates the variance issues.
However, this requires an accurate action-conditioned value function, which is notoriously hard to learn without relying on replay buffers for reusing past off-policy data.
We present Relative Entropy Pathwise Policy Optimization, an algorithm that trains Q-value models purely from on-policy trajectories, unlocking the use of Q function derivatives to compute policy updates in the context of on-policy learning.
We show how to combine stochastic policies for exploration with constrained updates for stable training, and evaluate important architectural components that stabilize value function learning.
This results in an efficient on-policy algorithm that combines the stability of Q-based policy gradients with the simplicity and minimal memory footprint of standard on-policy learning. Compared to state-of-the-art on two standard GPU-parallelized benchmarks, \algoname provides strong empirical performance at superior sample efficiency, wall-clock time, memory footprint, and hyperparameter robustness.

\end{abstract}

\begin{figure}[h]
    \centering
    \includegraphics[width=\linewidth]{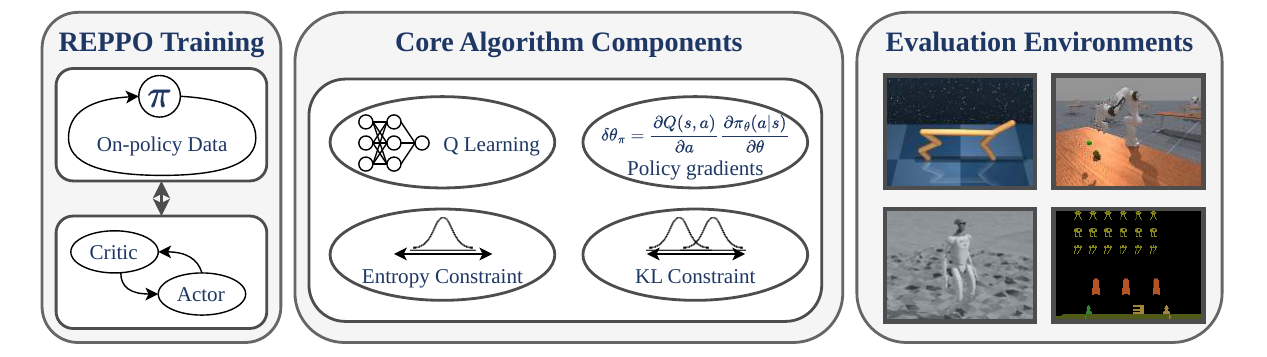}
    \caption{Overview of the \textbf{\blue{Relative Entropy Pathwise Policy Optimization}} algorithm. REPPO combines four core components to achieve stable and fast actor-critic training across a wide variety of reinforcement learning tasks. Isaac and Atari results can be found in the appendix. Implementations in different popular RL frameworks can be found at \url{https://github.com/reppo-rl}.}
    \label{fig:placeholder}
\end{figure}

\section{Introduction}

Most modern on-policy algorithms, such as TRPO \citep{schulman_trust_2015} or PPO \citep{schulman_proximal_2017}, use a score-based gradient estimator to update the policy.
These methods have proven useful for robotic control \citep{rudin2022learning, kaufmann2023champion, radosavovic2024real}, and language-model fine-tuning \citep{ouyang2022training, touvron2023llama, gao2023scaling, liu2024deepseek}, but are often plagued by training instability.
Zeroth-order, score-based gradient approximation exhibits high variance \citep{greensmith2004variance}, which leads to unstable learning \citep{ilyas202acloser, rahn2023policy}, especially in high-dimensional continuous spaces \citep{li2018policy}.
In addition, it requires importance sampling to allow sample reuse, which exacerbates the high variance.

Alternatively we can train a parameterized state-action value function \citep{lillicrap_continuous_2016,fujimoto_addressing_2018, haarnoja_soft_2018}, and use it to improve the policy, for example by using a \emph{pathwise} policy gradient \citep{silver_deterministic_2014}, i.e. taking the derivative of the Q function wrt the action.
Using a parameterized function to improve the policy often leads to faster and more stable learning learning by reducing the score-based estimators variance \citep{mohamed2020monte} and by allowing us to remove importance sampling corrections.

However, the effectiveness of these approaches is bounded by the quality of the approximate value function \citep{silver_deterministic_2014}.
As such, algorithms that use a state-action value function usually rely on improving value learning through off-policy training \citep{fujimoto_addressing_2018, haarnoja_soft_2018}.
Unfortunately, off-policy training requires the use of replay buffers.
Storing these replay buffers can be a challenge when the collected samples cannot fit in memory.
In addition, training with past data introduces various challenges for value function fitting \citep{thrun_issues_1993,baird_residual_1995,van_hasselt_double_2010, sutton_emphatic_2016,kumar_implicit_2021, nikishin_primacy_2022, lyle_disentangling_2024, hussing_dissecting_2024, voelcker_mad-td_2025}.
This raises our core question:

\begin{center}
\begin{minipage}{0.9\linewidth}
\centering
\emph{Can we train a sufficiently correct value function approximation and effectively use it for policy improvement in a fully on-policy setting without large replay buffers?}
\end{minipage}
\end{center}

\begin{figure}[t]
    \centering
    \includegraphics[width=\linewidth]{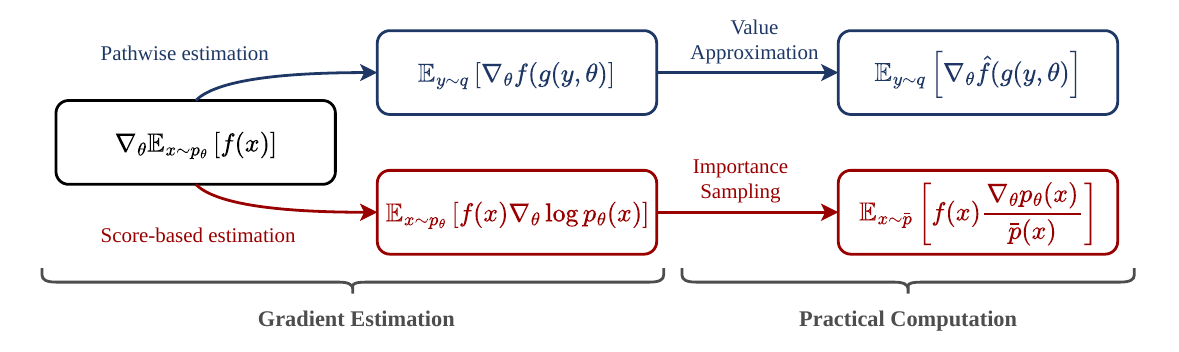}
    \caption{Overview of the strategies used by \blue{\algoname} and \red{PPO} to obtain policy gradient estimators. Computing the gradient requires a mathematical transformation that allows for efficient estimation from samples, and additional steps that make the computation tractable in practice.}
    \label{fig:overview}
\end{figure}

Building on the progress in accurate value function learning \citep{sutton_learning_1988, haarnoja_soft_2019, schwarzer_data-efficient_2021, hussing_dissecting_2024, farebrother_stop_2024}, we present an efficient on-policy algorithm, \emph{\fullalgoname~(\algoname)}, which uses the pathwise gradient estimator with an accurate value estimation trained on on-policy data.
\algoname builds on the maximum entropy framework \citep{ziebart2008maximum} to encourage exploration.
It combines this with a KL regularization scheme, inspired by the Relative Entropy Policy Search method \citep{peters_relative_2010}, which prevents aggressive policy updates from destabilizing the optimization.

Furthermore, we evaluate several prominent advances in neural network architecture design to stabilize learning: categorical Q-learning \citep{farebrother_stop_2024}, normalized neural network architectures \citep{nauman_overestimation_2024,hussing_dissecting_2024}, and auxiliary tasks \citep{jaderberg_reinforcement_2017}.
These components feature in many recent variants \citep{schwarzer_data-efficient_2021,schwarzer_bigger_2023,nauman_overestimation_2024,hussing_dissecting_2024,gallici_simplifying_2024,lee2025simba,lee2025hyperspherical,nauman2025bigger,fujimoto_towards_2024} of common value learning algorithm such as SAC \citep{haarnoja_soft_2018}.
We find that categorical Q-learning and normalization have a strong impact on the performance, while auxiliary tasks only show small impact, but become more relevant when reducing the amount of samples.

We test our approach in a variety of locomotion and manipulation environments from the Mujoco Playground \citep{zakka_mujoco_2025} and ManiSkill \citep{taomaniskill3} benchmarks, and show that \algoname is competitive with tuned on-policy baselines in terms of sample efficiency and wall-clock time, while using significantly smaller memory footprints than comparable off-policy algorithms.
Furthermore, we find that the proposed method is robust to the choice of hyperparameters.
To this end, our method offers stable performance across more than 30 tasks spanning multiple benchmarks with a single hyperparameter set.
In introducing \algoname, our work makes the following contributions:

\begin{enumerate}
    \item We showcase that using a state-action value function and a pathwise policy gradient can be effective in on-policy RL, as it allows on-policy action resampling, forgoing importance corrections.
    However, this requires learning a highly accurate state-action value function.
    \item We show how a joint entropy and policy deviation tuning objective can address the twin problems of sufficient exploration and controlled policy updates. 
    \item We evaluate architectural components such as cross-entropy losses, layer normalization, and auxiliary tasks for their efficacy in pathwise policy gradient-based on-policy learning.
\end{enumerate}

We provide sample implementations in both the \lstinline{JAX} \citep{jax2018github} and \lstinline{PyTorch} \citep{paszke_pytorch_2019} frameworks at \url{https://github.com/reppo-rl}.

\section{Background, notation, and definitions}

We consider the setting of the Markov Decision Process (MDP)~\citep{puterman1994markov} , defined by the tuple $(\mathcal{X}, \mathcal{A}, \mathcal{P}, r, \gamma, \rho_0)$, where $\mathcal{X}$ is the set of states, $\mathcal{A}$ is the set of actions, $\mathcal{P}(x'|x, a)$ is the transition probability kernel, $r(x, a)$ is the reward function, and $\gamma \in [0,1)$ is the discount factor. We write $\mathcal{P}_\pi(x'|x)$ for the policy-conditioned transition kernel and $\mathcal{P}_\pi^n(y|x)$ for the n-step transition kernel.
An agent interacts with the environment via a policy $\pi(a | x)$, which defines a distribution over actions given a state. The objective is to find a policy that maximizes the expected discounted return,
$J(\pi) = \mathbb{E}_{\pi} \left[ \sum_{t=0}^{\infty} \gamma^t r(x_t, a_t) \right],$
where $x_0 \sim \rho_0$ is the initial state distribution, and $a_t \sim \pi(\cdot | x_t)$.
The state-action value function associated with a policy $\pi$ are defined as
$Q^\pi(x, a) = \mathbb{E}_{\pi} \left[ \sum_{t=0}^\infty \gamma^t r(x_t, a_t) \Big| x_0 = x, a_0 = a \right].$
We use $\mu_\pi(y|x)$ to denote the discounted stationary distribution over states $y$ when starting in state $x$.
When $x \sim \mu_\pi(\cdot|y)$, $y \sim \rho_0$, we will simply write $\mu_\pi(x)$ to denote the probability of a state under the discounted occupancy distribution.\footnote{Many policy gradient methods use the undiscounted empirical state occupancy for optimization \citep{nota202is}. While this could be considered an error, it is nonetheless a common approximation. \algoname similarly uses empirical samples without accounting for the discount factor in the objective.}

\subsection{Policy gradient learning}
\label{sec:pg}

A policy gradient approach~\citep{sutton_reinforcement_2018} is a general method for improving a (parameterized) policy $\pi_\theta$ by estimating the gradient of the policy-return function $J(\pi_\theta)$ with regard to the policy parameters $\theta$.
The \emph{policy gradient theorem} states that
\begin{align}
    \nabla_\theta J(\pi_\theta) = \mathbb{E}_{x \sim \mu_\pi, a\sim\pi_\theta(\cdot|x)}[Q^{\pi_\theta}(x,a) \nabla_\theta \log \pi_\theta(a|x)]. \label{eq:mcpg}
\end{align}
This identity is particularly useful as both the Q value and the stationary distribution can be estimated by samples obtained from following the policy for sufficiently many steps in the environment.

An alternative approach is the \emph{deterministic policy gradient theorem} (DPG) \citep{silver_deterministic_2014}.
The estimator for the DPG relies on access to a differentiable state-action value function and a deterministic differentiable policy $\pi^\mathrm{det}_\theta(x)$.
While access to the true value function is an unrealistic assumption, we can use a trained surrogate model, $\hat{Q}$, to obtain a biased estimate of the gradient
\begin{align}
    \nabla_\theta J(\pi_\theta) \approx \mathbb{E}_{x \sim \mu_\pi}[\nabla_a \hat{Q}^{\pi^\mathrm{det}_\theta}(x,a)|_{a = \pi^\mathrm{det}_\theta(x)} \nabla_\theta \pi^\mathrm{det}_\theta(x)]. \label{eq:dpg}
\end{align}

Finally, the DPG can be expanded to reparametrizable stochastic policies\footnote{We discuss an extension to non-reparametrizable, discrete policies in \autoref{app:dreppo}.}.
We term this the \emph{pathwise policy gradient}, following \cite{mohamed2020monte}, but the formulation has been used prominently in prior work such as SAC \citep{haarnoja_soft_2018}, just without a proper name.
The gradient estimator can be obtained from the following expectation
\begin{align}
    \nabla_\theta J(\pi_\theta) \approx \mathbb{E}_{x \sim \mu_\pi, \epsilon\sim p(\epsilon)}[\nabla_a \hat{Q}^{\pi^\mathrm{rep}_\theta}(x,a)|_{a = \pi^\mathrm{rep}_\theta(x,\epsilon)} \nabla_\theta \pi^\mathrm{rep}_\theta(x, \epsilon)], \label{eq:ppg}
\end{align}
where $\pi^\mathrm{rep}_\theta(x,\epsilon)$ is a reparameterization of $\pi_\theta(a|x)$.
To avoid notational clutter we will write $\pi_\theta(a|x)$ from now on to always mean the appropriate reparameterization.

\begin{figure}
    \begin{subfigure}{\linewidth}
        \centering
        \begin{minipage}{0.49\linewidth}
            \includegraphics[width=0.9\linewidth]{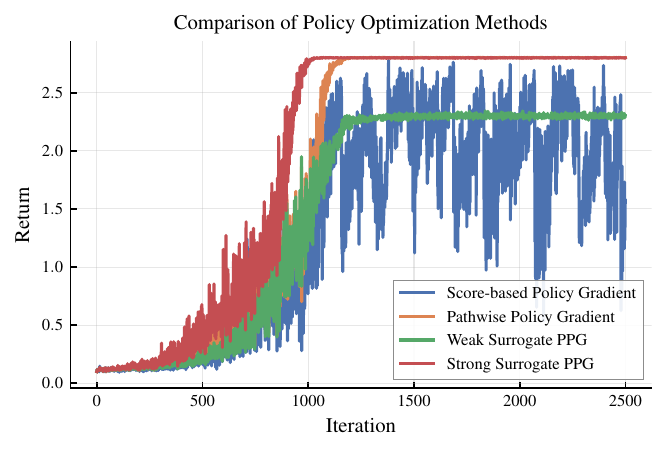}
        \end{minipage}
        \hfill
        \begin{minipage}{0.49\linewidth}
            \includegraphics[width=0.9\linewidth]{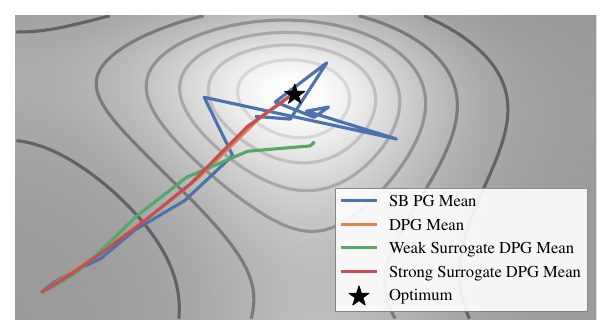}
        \end{minipage}
        \caption{Achieved returns (left) and path of four policies trained with different gradient estimation methods. We compare a score-function based policy gradient estimator (blue) with three variants of pathwise gradient estimators: using the ground truth objective function (orange), an inaccurate surrogate model (green), and an accurate surrogate model (red). All PPG based methods show markedly reduced variance in the policy updates.}
        \label{fig:pg_comparison}
    \end{subfigure}
    
    \begin{subfigure}{\linewidth}
        \centering
        \includegraphics[width=0.98\linewidth]{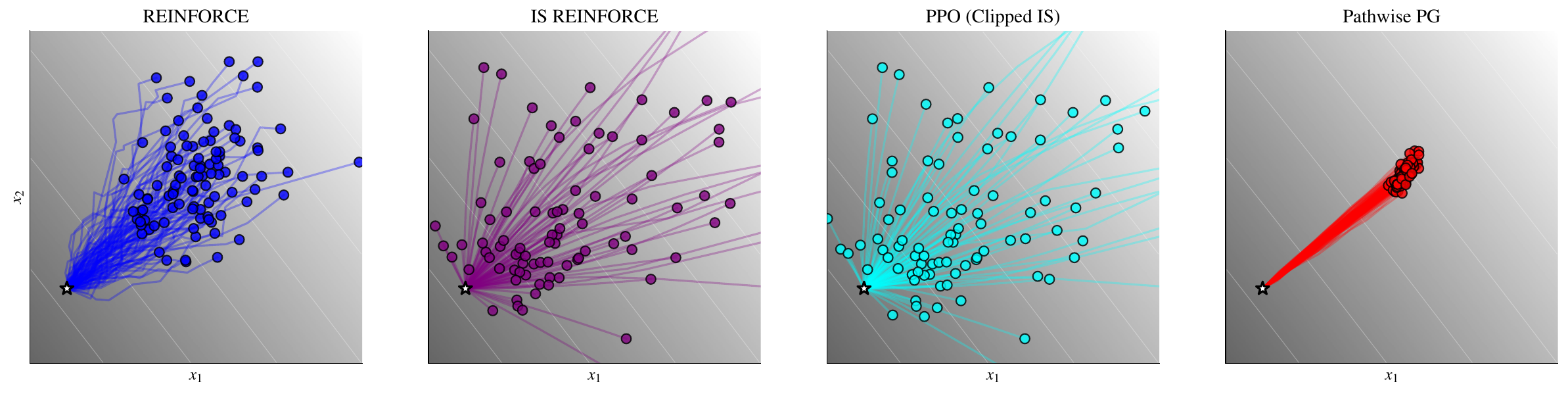}
        \caption{Gradient path over eight steps in the middle of the trajectory, visualized per algorithm for 8 steps. For Reinforce and PPG, new samples are drawn at every step. For the importance sampling based algorithms, one set of samples is sampled at the beginning and subsequent steps are conducted using importance sampling.}
        \label{fig:pg_varaince}
    \end{subfigure}
    \caption{Visualization of gradient paths on a 2D example function.}
\end{figure}
\subsection{Understanding sources of harmful variance in gradient estimation}
\label{sec:toy}
To build additional intuition on the differences between different policy gradient estimators, we conduct an illustrative experiment.
Implementation details can be found in \autoref{app:example}.

We initialize four Gaussians and update their parameters to maximize
    $J(\mu, \Sigma) = \mathbb{E}_{x \sim \mathcal{N}(\cdot|\mu,\Sigma)}[g(x)]$
on a test function $g(x)$ with four different methods: a score-based policy gradient (using \autoref{eq:mcpg}), a pathwise PG with the ground truth objective function, and two pathwise PGs using learned approximations, one accurate and one inaccurate (all using \autoref{eq:ppg}).
We visualize the returns and the path of the mean estimates in \autoref{fig:pg_comparison}.
In addition, we zoom in on the gradient paths of the score-based estimator.
We visualize 100 different eight step paths from the middle of the trajectory.
Here, in addition to the vanilla score-based estimator, we also show an importance sampling and a clipped importance sampling estimator.
These paths are visualized in \autoref{fig:pg_varaince}.

The experiment shows that score-based gradient estimators have high variance, and can lead to unstable policies which fail to optimize the target.
In addition, while importance sampling increases the sample efficiency of the algorithm, it greatly exacerbates these variance issues.
We find that clipping the ratio estimate, as proposed by \cite{schulman_proximal_2017}, prevents catastrophic instability, but does not reduce the variance substantially.
On the other hand, using a pathwise gradients is remarkably stable and exhibits small variance.
However, it either requires access to the gradients of the objective function, or a strong surrogate model.

To use pathwise gradients in on-policy learning, our goal is thus to learn a suitable value function that allows us to estimate a low variance update direction without converging to a suboptimal solution.

\section{Relative Entropy Pathwise Policy Optimization}
Naively, we could take an off-policy algorithm like SAC and train it solely with data from the current policy.
However, as \cite{seo2025fasttd3} recently showed, this can quickly lead to unstable learning. 
To succeed in the on-policy regime, we need to be able to continually obtain new diverse data, and compute stable and reliable updates. 
Combining a set of recent advances in both reinforcement learning as well as neural network value function fitting, can satisfy these requirements. 
We first introduce the core RL algorithm, and then elaborate on the architectural design of the method. 

At its core, \algoname proceeds similar to other on-policy actor-critic algorithms through three distinct phases: data gathering, value target estimation, and value and policy learning (see \autoref{alg:op-sac}). To obtain diverse data, \algoname uses a maximum-entropy formulation, adapted to multi-step TD-$\lambda$ (\autoref{sec:sac}), to encourage exploration.
Finally, to ensure that policies do not collapse and policy learning is stable, \algoname uses KL-constrained policy updates with a schedule that balances entropy-driven exploration and policy constraints (\autoref{sec:kl}). 

\subsection{Value function learning}
\label{sec:sac}

Off-policy PPG methods like TD3 \citep{fujimoto_addressing_2018} and SAC \citep{haarnoja_soft_2018} mostly use single step Q learning, i.e. they use only immediate rewards for value function updates. This is paired with large replay buffers to stabilize learning.
While on-policy algorithms cannot use past policy data, they can instead use low bias multi-step TD targets for stabilization \citep{fedus_revisiting_2020}.
Therefore, multi-step TD-$\lambda$ targets form the basis for our value learning objective.
Note that REPPO is more closely related to SARSA than to Q-learning \citep{sutton_reinforcement_2018}, due to being on-policy.

In addition to multi-step returns, diverse data is crucial.
To achieve a constant rate of exploration, and prevent the policy from prematurely collapsing to a deterministic function, we leverage the maximum entropy formulation for RL \citep{ziebart2008maximum,levine_reinforcement_2018}.
The core aim of the maximum entropy framework is to keep the policy sufficiently stochastic by solving a modified policy objective which not only maximizes rewards but also penalizes the loss of entropy in the policy distribution.
The maximum-entropy policy objective \citep{levine_reinforcement_2018} can be defined as
\begin{align}
    J_\text{ME}(\pi_\theta) = \mathbb{E}_{\pi_\theta} \left[ \sum_{t=0}^{\infty} \gamma^t r(x_t, a_t) + \alpha \mathcal{H}[\pi_\theta(x_t)]\right],
\end{align}
where $\mathcal{H}[\pi_\theta(x)]$ is the entropy of the policy evaluated at $x$, and $\alpha$ is a hyperparameter which trades off reward maximization and entropy maximization.
\algoname combines the maximum entropy objective with TD-$\lambda$ estimates, resulting in the following target estimate 
\begin{align}
    G^{(n)}(x_t,a_t) &= \sum_{k=t}^{n} \gamma^{k-t} (r(x_k,a_k) - \alpha \log \pi(a_k|x_k)) + \gamma^{n+1} Q(x_n,a_n)\\
    G^\lambda(x,a) &= \frac{1}{\sum_{n=0}^N\lambda^n} \sum_{n=0}^N \lambda^n G^{(n)}(x,a),
\end{align}
where $N$ is the maximum length of the future trajectory we obtain from the environment for the state-action pair $(x,a)$.
Our implementation relies on the efficient backwards pass algorithm presented by \cite{daley_2019_reconciling}.
Crucially, the targets are computed on-policy after a new data batch is gathered, and the Q targets are not recomputed before gathering new data.
Our Q learning loss is
\begin{align}
    \mathcal{L}^\mathrm{\algoname}_Q\left(\phi|\{x_i,a_i\}_{i=1}^{B}\right) &= \frac{1}{B} \sum_{i=1}^{B} \mathrm{HL}\left[Q_\phi(x_i, a_i), G^{\lambda}(x_i,a_i)\right] + \mathcal{L}_\mathrm{aux}(f_\phi(x_i, a_i), x'_i),
\end{align}
where $x'_i$ refers to the next state sample starting from $x_i$, and $\mathrm{HL}$ is the HL-Gauss loss (see \autoref{sec:gauss} and \autoref{app:gauss}), and $\mathcal{L}_\mathrm{aux}$ is presented in \autoref{sec:aux} and \autoref{app:aux_task}.

Using purely on-policy targets allows us to remove several common off-policy stabilization components from the value learning setup.
\algoname does not require a pessimism bias, so we can forgo the clipped double Q learning employed by many prior methods~\citep{fujimoto_addressing_2018}.
Tuning pessimistic updates carefully to allow for exploration is a difficult task \citep{moskovitz_tactical_2021}, so this simplification increases the robustness of our method.
We also do not need a target value function copy, since we do not recompute the target at each step and it therefore remains on-policy.

\subsection{Policy Learning}
\label{sec:kl}

A core problem with value-based on-policy optimization is controlling the size of the policy update, as the value estimate is only accurate on the data covered by the prior policy.
A large policy update can therefore destabilize learning \citep{kakade2002approximately}.
This problem has led to the development of constrained policy update schemes, where the updated policy is prevented from deviating too much from the behavioral
\citep{peters_relative_2010,schulman_trust_2015}.
To control the deviation, we use the Kullback-Leibler (KL) divergence, also called the relative entropy \citep{peters_relative_2010}, as it can be justified theoretically through information geometry \citep{kakade2001natural,peters_natural_2008,pajarinen2019compatible}, and is easy to approximate using samples.

Some works in the literature \citep{neumann2011variational,sokota2022a} claim that the reverse mode might be preferable for policy constraints, as it is mode-seeking, and the forward mode is mode-averaging.
However, this intuition does not cleanly translate to our setting.
As our policies are unimodal tanh-squashed Gaussian, the main impact of the KL direction is that the reverse-mode KL is entropy reducing.
As we explicitly aim to increase the policy's entropy using the maximum entropy formulation, using forward-mode KL makes the optimization more stable.

\paragraph{Policy Optimization Objective}
Our policy updates derive from a constrained optimization problem which includes both entropy and the KL constraint, and where $\theta'$ is the behavior policy, and $\varepsilon_\mathrm{KL}$ and $\varepsilon_\mathcal{H}$ are the respective KL and entropy constraints
\begin{align}
\max_{\theta} \quad & \mathbb{E}_{x\sim\rho_{\pi_{\theta'}}} \left[ \mathbb{E}_{a\sim\pi_\theta(\cdot|x)}\left[Q(x,a)\right]\right] \\
\text{subject to} \quad &  \mathbb{E}_{x\sim \rho_{\pi_{\theta'}}}\left[D_{\mathrm{KL}} \left( \pi_{\theta'}(\cdot|x) \, \| \, \pi_{\theta}(\cdot|x) \right) \right] \leq \varepsilon_\mathrm{KL}\\
& \mathbb{E}_{x\sim \rho_{\pi_{\theta'}}} \left[\mathcal{H}[\pi_{\theta}(\cdot|x)] \right] \geq \varepsilon_\mathcal{H}.
\end{align}
A similar combination of maximum entropy and KL divergence bound has been explored in various forms \citep{abdolmaleki2015model,pajarinen2019compatible,akrour2019projections}.
However, while previous approaches use complex solutions to this problem, such as approximate mirror descent, line search, or heuristic clipping, we take a simpler approach.
We relax the problem, which introduces two hyperparameters, $\alpha$ for the entropy, and $\beta$ for the KL.
Inspired by \cite{haarnoja_soft_2019}, \algoname automatically adapts these constraints when the policy violates them.

\paragraph{Policy Updates and Multiplier Tuning}
\label{sec:ent_tuning}
In the constrained objective, we introduce two hyperparameters, $\varepsilon_\mathcal{H}$ and $\varepsilon_\mathrm{KL}$, which bound the entropy and KL divergence.
The goal of the Lagrangian parameters is to ensure that the policy stays close to these constraints.
As we need to ensure that they remain positive, we update them in log space with a gradient based root finding procedure 
\begin{align}
    \alpha &\leftarrow \alpha - \eta_\alpha e^\alpha \mathbb{E}_{x \sim \rho_{\pi_{\theta'}}}\left[(\mathcal{H}[\pi_\theta(\cdot|x)] - \varepsilon_\mathcal{H})\right]\\
    \beta  &\leftarrow \beta - \eta_\beta  e^\beta \mathbb{E}_{x \sim \rho_{\pi_{\theta'}}}\left[(\mathrm{D}_\mathrm{KL}(\pi_{\theta'}(\cdot|x) \| \pi_{\theta}(\cdot|x))- \varepsilon_\mathrm{KL})\right].
\end{align}

Finally, to ensure our KL constraint is (approximately) maintained, we clip the actor loss based on whether the constrained is currently violated.
The full policy objective for \algoname is now
\begin{align}
    \mathcal{L}^\mathrm{REPPO}_\pi(\theta|x_i) &= 
    \begin{cases}
    -Q(x_i,a) + e^\alpha \log \pi_\theta(a|x_i),&\text{if }\frac{1}{k}\sum_{j=1}^k \log \frac{\pi_{\theta'}(a_j|x_i)}{\pi_\theta(a_j|x_i)} < \varepsilon_\mathrm{KL} \\
    e^\beta \frac{1}{k}\sum_{j=1}^k \log \frac{\pi_{\theta'}(a_j|x_i)}{\pi_\theta(a_j|x_i)},&\text{otherwise}
\end{cases}
\label{eq:reppo_pi}
\end{align}
where $a$ is sampled from $\pi_\theta(\cdot|x_i)$ and $a_j$ from the past behavior policy $\pi_{\theta'}(\cdot|x_i)$, and $k$ denotes how many samples are used to approximate the KL.
As with the critic, the optimized loss is a mean over a minibatch from the rollout data. 
Note that contrary to other on-policy algorithms like PPO and TRPO, we are not forced to use actions sampled from the behavior policy in the policy gradient estimator, which removes the need for importance sampling correction.
We will show that this greatly improves the performance of REPPO in \autoref{sec:score}.

Jointly tuning the entropy and KL multipliers is a crucial component of \algoname.
As the policy entropy and KL are tied, letting the entropy of the behavior policy collapse means the KL constraint prevents any policy updates.
Furthermore, the entropy and KL terms are balanced against the scale of the returns in the maximum entropy formulation.
As the returns increase, keeping the multipliers fixed causes the model to ignore the constraints over time, accelerating collapse.
However, as we tune both in tandem, our setup ensures a steady, measured update rate of the policy.

\subsection{Stable Representation and Value Function Architectures}

While the RL algorithm offers a strong foundation to obtain strong surrogate values, we also draw on recent off-policy advances in value function learning that improve training through architecture and loss design.
We incorporate three major advancements into \algoname to further stabilize training.

\paragraph{Cross-entropy loss for regression}
\label{sec:gauss}
The first choice is to replace the mean squared error in the critic update with a more robust cross-entropy based loss function.
For this, \algoname uses the HL-Gauss loss \citep{farebrother_stop_2024}.
This technique was adapted from the distributional C51 algorithm \citep{bellemare_distributional_2017}, which can lead to stable learning algorithms.
Inspired by this insight and histogram losses for regression~\citep{imani2018improving}, \cite{farebrother_stop_2024} hypothesize that observed benefits are due to the fact that many distributional algorithms use a cross-entropy loss, which is scale invariant.
\cite{palenicek2025xqc} further investigate and reinforce this claim, showing that stable gradients arise from cross-entropy based losses.
We present the mathematical form of the loss formulation in \autoref{app:gauss}.
We find that a categorical loss is a crucial addition, as our ablation experiments show (\autoref{sec:ablations}), but alternatives like C51 could easily work as well.

\paragraph{Layer Normalization}
\label{sec:norm}
Several recent works \citep{ball_efficient_2023, yue2023understanding, lyle_disentangling_2024, nauman_overestimation_2024, hussing_dissecting_2024,gallici_simplifying_2024} have shown the importance of layer normalization \citep{ba_layer_2016} for stable critic learning.
\cite{gallici_simplifying_2024} provides a thorough theoretical analysis of the importance of normalization in on-policy learning, while \cite{hussing_dissecting_2024} focuses on assessing the empirical behavior of networks in off-policy learning with and without normalization.
As we operate in an on-policy regime where value function targets are more stable, we find that normalization is not as critical for \algoname as it is for off-policy bootstrapped methods; yet, we still see performance benefits in most environments from normalization.

\paragraph{Auxiliary tasks}
\label{sec:aux}
Auxiliary tasks \citep{jaderberg_reinforcement_2017} can stabilize features in environments with sparse rewards, where the lack of a reward signal can prevent learning meaningful representations via the Q learning objective \citep{voelcker_can_2024}.
For REPPO, auxiliary tasks are especially impactful when we decrease the number of samples used in each update batch (see \autoref{sec:ablations}).
We provide a discussion of this auxiliary task setup, including the loss function, in \autoref{app:aux_task}.

\section{Experimental Evaluation}
\label{sec:experimental}

We begin by evaluating whether pathwise estimators improve upon score-based estimation in on-policy RL settings. We then compare our approach to baselines, evaluating final performance, sample and wall-clock efficiency, and stability of policy improvement. Our results demonstrate strong performance of \algoname on all axes. Additional details on architectures, hyperparameters, and ablations are provided in \autoref{app:impl} and \autoref{app:experiments}. A discrete variant of REPPO, along with its architectural changes and experimental results, is presented in \autoref{app:dreppo}.

\begin{figure}[b]
    \centering
    \includegraphics[width=.95\linewidth]{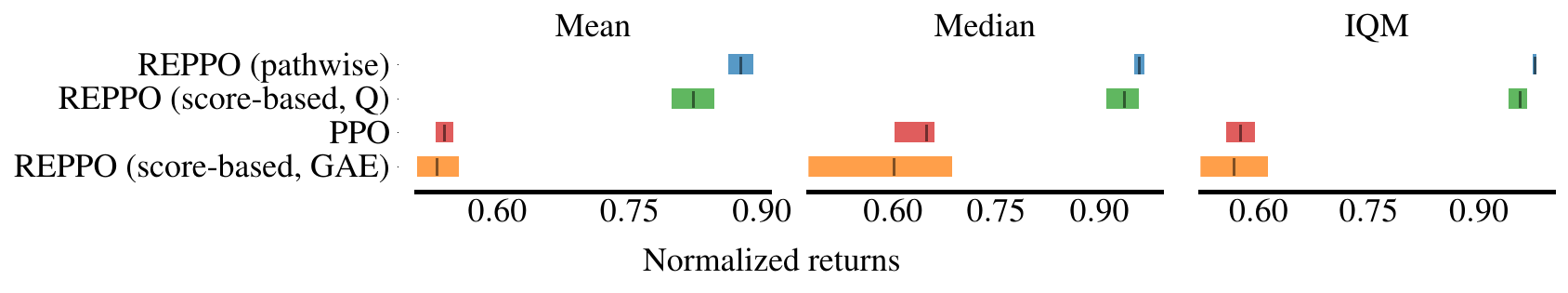}
    \caption{\textbf{Aggregate performance \lstinline{mujoco_playground}} We compare \algoname with two ablations: using the score-based gradient estimator with the learned Q function, and using an on-policy GAE estimate with importance sampling and clipping. For additional context, we also report PPO results.}
    \label{fig:score_agg}
\end{figure}

\paragraph{Environments}
We evaluate \algoname on two major GPU-parallelized benchmark suites: 23 tasks from the \lstinline{mujoco_playground} DMC suite \citep{zakka_mujoco_2025} and 8 ManiSkill environments \citep{taomaniskill3}, covering locomotion and manipulation, respectively. 
These tasks span high-dimensional control, sparse rewards, and chaotic dynamics.

\subsection{Score-based and Pathwise Comparison}
\label{sec:score}

REPPO offers an alternative to score-based policy gradient estimation in on-policy RL. However, we also introduce several enhancements, including automated tuning of entropy and KL coefficients, to improve value and policy learning. To assess the benefits of learned values and pathwise gradient estimation over score-based methods, we conduct two experiments. First, we replace the pathwise term $-Q(x,a)$ in \autoref{eq:reppo_pi} with the score function $\log \pi(a|x) [Q(x,a)]_\mathrm{sg}$, denoted as \emph{REPPO (score-based, Q)}. Second, we replace the gradient estimator with the GAE-based clipped objective from PPO, denoted as \emph{REPPO (score-based, GAE)}.
Aggregate results are presented in \autoref{fig:score_agg}.

Using the approximate Q function in the policy gradient objective provides a strong improvement over a clipped objective.
This showcases the benefits of value function learning and removing importance sampling.
This also shows that the \algoname framework can be used with policy classes that are not amenable to reparameterization, such as diffusion policies \citep{chi2024diffusionpolicy,celik2025dime,ma2025efficient}, by using a score-based estimator together with the learned Q function.
Interestingly, combining the PPO objective with REPPO leads to slightly worse results than vanilla PPO. 
We find that the high variance complicates the automatic parameter tuning scheme.

\subsection{Benchmark comparison}
\begin{figure}
    \centering

    \begin{subfigure}{\linewidth}
        \centering
        \includegraphics[width=\linewidth]{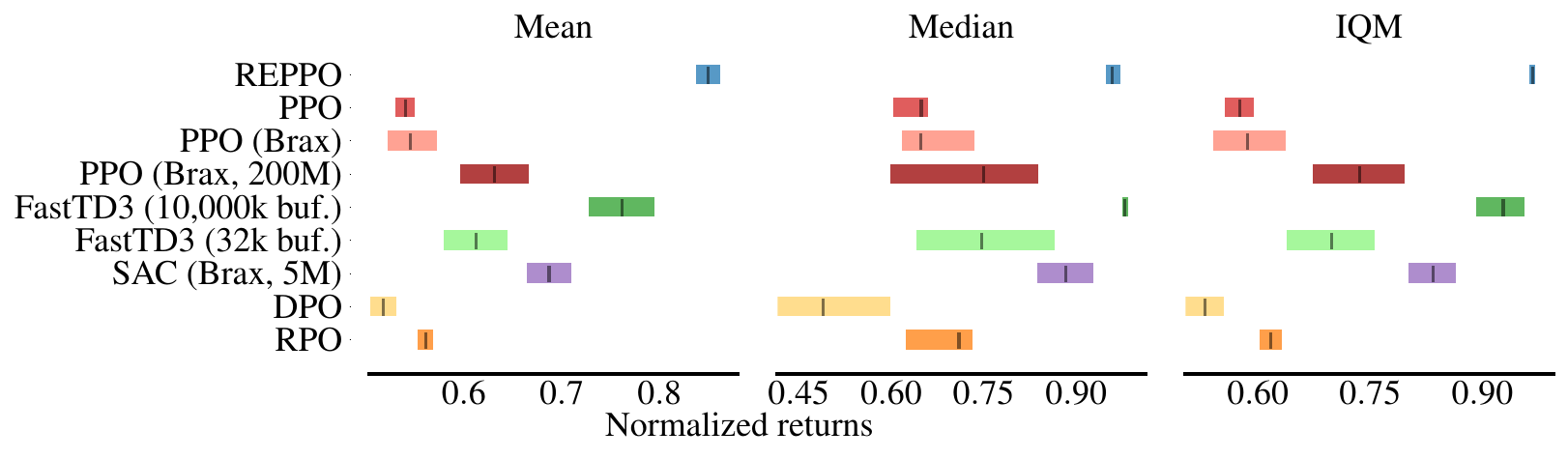}
        \caption{\textbf{Aggregate performance \lstinline{mujoco_playground}}. We compare \algoname to a re-tuned PPO baseline, the Brax PPO and SAC implementations provided by \cite{zakka_mujoco_2025}, as well as FastTD3 \citep{seo2025fasttd3}, RPO \citep{rahman2023robust}, and DPO \citep{lu2022discovered}.
        }
        \label{fig:rliable_overview_dmc}
    \end{subfigure}

    \vspace{0.8em}  %

    \begin{subfigure}{\linewidth}
        \raggedleft
        \includegraphics[width=0.905\linewidth]{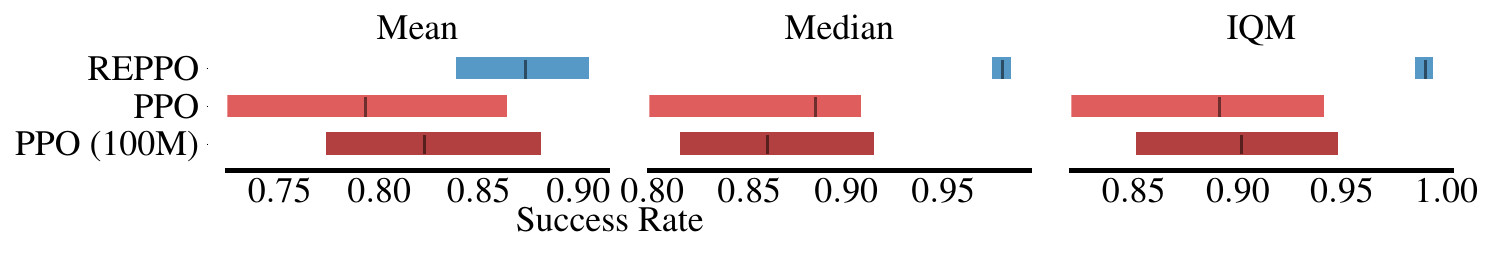}
        \caption{\textbf{Aggregate success \lstinline{maniskill} \citep{taomaniskill3}}. We compare \algoname against a PPO baseline provided by \cite{taomaniskill3} at 50 million environment steps. As some environments take more than 50 million steps for PPO to achieve strong performance, we report the final performance at 100 million steps. While the mean confidence intervals are very broad, \algoname performs strongly on the IQM and median metrics.}
    \end{subfigure}

    \caption{Aggregate performance comparison on  (a) \lstinline{mujoco_playground} DMC  and (b) ManiSkill3.}%
    \label{fig:rliable_overview}
\end{figure}

We compare \algoname against the PPO and SAC results reported by \citet{zakka_mujoco_2025} and \citet{taomaniskill3}. 
We report PPO baselines at 50M environment steps, and at the larger training horizon used in the original papers \citep{zakka_mujoco_2025}.
Results taken from \cite{zakka_mujoco_2025} are denoted as ``PPO/SAC (Brax)''.
To ensure that PPO is not undertuned for the 50m step regime we re-tuned the hyperparameters of the implementation provided by \cite{lu2022discovered}.
SAC results are reported at 5m steps as this amounts to  similar total runtime as the 200m PPO results (compare results in \cite{zakka_mujoco_2025}.
Naively running SAC at a larger sample budget and wall-clock efficiency can lead to instability, as \cite{seo2025fasttd3} demonstrates.
Furthermore, we include FastTD3~\citep{seo2025fasttd3} on DMC locomotion tasks, trained under two memory budgets: the default replay buffer (10,485,760 transitions) and a constrained buffer similar in size to on-policy methods (32,768 transitions) to control for the the memory and performance trade-off. 
Finally, we compare against Robust Policy Optimization (RPO) \citep{rahman2023robust} and Discovered Policy Optimization (DPO) \citep{lu2022discovered}.
However, even with some hyperparameter tuning, we were unable to achieve a strong performance improvement beyond the PPO baseline with these approaches.

For \algoname, we report results aggregated over 20 seeds.
We run 20 seeds for PPO and 5 for FastTD3\footnote{We use fewer seeds for FastTD3 as we are unable to replicate the speed claimed in the paper. This is due pytorch specific issues discussed in \autoref{app:wallclock}, and because we use smaller GPUs for our experiments.}, reporting aggregate scores with 95\% bootstrapped confidence intervals~\citep{agarwal_deep_2021}. To aggregate scores, \lstinline{mujoco_playground} returns are normalized by the maximum achieved by any algorithm.
For ManiSkill we report raw success rates, which are comparable across tasks.

\begin{figure}
    \centering
    \begin{minipage}{0.49\textwidth}
    \centering
    \includegraphics[width=1.0\linewidth]{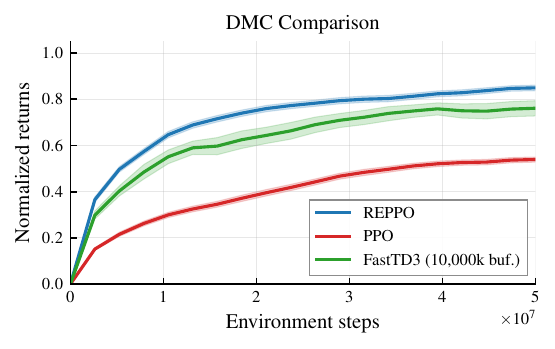}
    \end{minipage}
    ~
    \begin{minipage}{0.49\textwidth}
    \centering
    \includegraphics[width=1.0\linewidth]{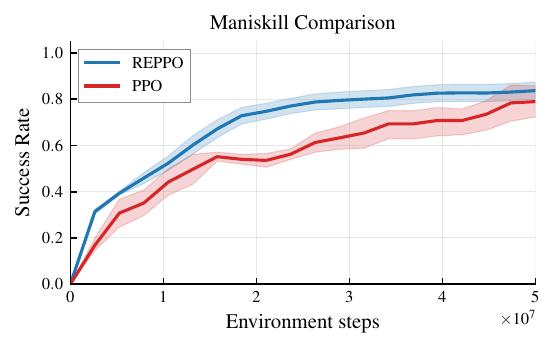}
    \end{minipage}
    \caption{Aggregate sample efficiency curves for the benchmark environments. Settings are identical to those in \autoref{fig:rliable_overview}. \algoname achieves higher performance at a faster rate in both benchmarks.}
    \label{fig:sample_efficiency}
\end{figure}

\paragraph{Final Performance and Sample Efficiency}
We first investigate the performance of policies trained using \algoname.
We report aggregate performance at the end of training on both benchmarks in \autoref{fig:rliable_overview}. 
For both benchmarks, we also provide the corresponding training curves in \autoref{fig:sample_efficiency}.

The aggregate results shown in \autoref{fig:rliable_overview} and \autoref{fig:sample_efficiency} indicate that our proposed method achieves statistically significant performance improvements over PPO, as well as similar performance to FastTD3 despite \algoname being fully on-policy. Although these results are most pronounced in locomotion tasks, ManiSkill manipulation results show significant performance benefits over PPO in terms of outlier-robust metrics~\citep{chan_measuring_2020,agarwal_deep_2021}. 

We find that PPO struggles on high-dimensional tasks such as HumanoidRun. Moreover, despite its approximate trust-region updates, PPO suffers from performance drops and unstable training. This erratic behavior closely mirrors the score-based policy gradient instability shown in \autoref{fig:pg_comparison}. In contrast, \algoname exhibits more stable improvements and lower variance across seeds.

\begin{wrapfigure}{r}{0.45\linewidth} 
  \includegraphics[width=1.0\linewidth]{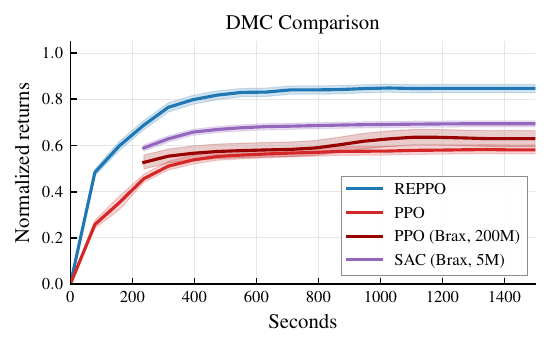}
  \caption{Wall-clock time comparison of \algoname against PPO and SAC implementations in JAX. \algoname matches other algorithms' speed but achieves higher return.}
  \label{fig:wallclock}
  \vspace{-10pt}
\end{wrapfigure}
\paragraph{Wall-clock Time}
Wall-clock time is an important metric, as it reflects the practical utility of an algorithm: faster training enables more efficient hyperparameter search and experimentation. However, measuring wall-clock time is nuanced, as results heavily depend on implementation details and are difficult to reproduce. We discuss these challenges across different frameworks in \autoref{app:wallclock}. In \autoref{fig:wallclock}, we compare the wall-clock performance of our approach against PPO and SAC in JAX. Other baselines lack JIT-compilable implementations, making direct comparisons less meaningful.

The computational cost per update is higher for \algoname than for PPO due to larger default networks and gradient propagation through the critic–actor chain. Nevertheless, both algorithms converge on most tasks in roughly 600–800 seconds, with \algoname achieving about 33\% higher normalized returns. This shows that the sample efficiency of pathwise gradients can offset their higher per-update cost, yielding improved wall-clock efficiency compared to score-based PPO.
In addition, we find that jax-based SAC, which is tuned to trade sample for computational efficiency, slightly outperforms PPO, but does not match REPPO in performance.
We note that other, modern SAC implementations \citep{nauman_bigger_2024,lee2025simba,lee2025hyperspherical}, are able to achieve better performance, but at the cost of computational efficiency.

\begin{figure}
    \centering
    \begin{minipage}{0.4\linewidth}
    \centering
    \includegraphics[width=1.0\linewidth]{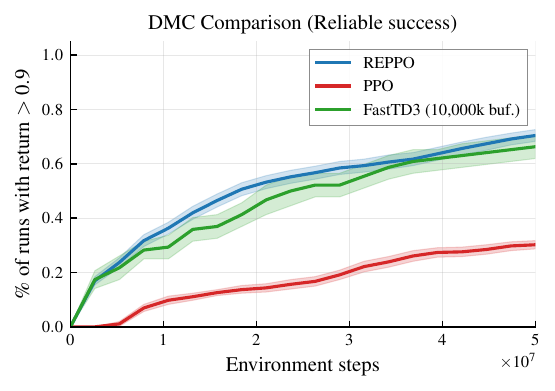}
    \end{minipage}
    ~~
    \begin{minipage}{0.4\linewidth}
    \centering
    \includegraphics[width=1.0\linewidth]{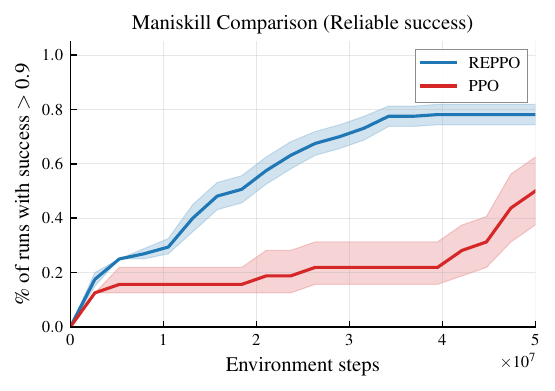}
    \end{minipage}
    \caption{Fraction of runs that achieve reliable performance as measured by our metric for policy stability and reliability. \algoname's immediately starts achieving high performance in some runs and the number gradually increases indicating stable learning. PPO struggles to achieve high performance initially and to maintain high performance throughout training.}
    \label{fig:reliable_maniskill}
\end{figure}

\paragraph{Reliable Policy Success}
We further investigate the stability of policy improvements using score-based and pathwise policy gradients. 
Our guiding principle is that such updates should not cause large drops in performance. 
To capture this, we adopt the ``reliable success'' metric, as proposed in \citet{chan2020measuring}.
We define an algorithm as \emph{reliably performant} if, once its performance exceeds a fixed threshold $\tau$, it never drops below this threshold thereafter. 
At each timestep, we track the number of runs that satisfy this criterion. 
This metric reflects the practical requirement that a deployed algorithm should not suddenly degrade simply due to continued training. 
We report the percentage of reliably successful runs for both \algoname and PPO in \autoref{fig:reliable_maniskill}.

On both DMC and ManiSkill benchmarks, \algoname achieves reliable performance improvements quickly. By the end of training, about four out of five runs have reached the threshold of $\tau=0.9$, whereas PPO achieves roughly 40 percentage points fewer reliably performant runs. We also find notable differences in sample efficiency: PPO requires 5–10 million interactions before most envs become reliably performant. Overall, these results show that, despite relying on a biased value model, pathwise policy gradients enable stable long-term improvement.

\section{Conclusion and avenues for future work}
\label{sec:conc}

In this paper we present \algoname, a highly performant yet efficient on-policy algorithm that leverages trained state-action value functions and pathwise policy gradients. By balancing entropic exploration and KL-constraints, and incorporating recent advances in neural network value function learning, \algoname is able to learn a high-quality approximation sufficient for reliable gradient estimation. As a result, the algorithm outperforms PPO on two GPU-parallelized benchmarks in terms of final return, sample efficiency and reliability while being on par in terms of wall-clock time. In addition, the algorithm does not require storing large amount of data making it competitive with recent advances in off-policy RL while requiring orders of magnitude lower amounts of memory.

As our method opens a new area for algorithmic development, it leaves open many exciting avenues for future work.
As \cite{seo2025fasttd3} shows, using replay buffers can be beneficial to stabilize learning as well.
This raises the question if our Q learning objective can be expanded to use both on- and off-policy data to maximize performance while minimizing memory requirements.
Furthermore, the wide literature on improvements on PPO, such as learned objectives \citep{lu2022discovered} can be incorporated into \algoname.
We also observe that removing the importance sampling step in PPO has a crucial impact on performance, which suggests further research on the trade-off between efficiency and stability in on-policy gradient estimation is needed.
Finally, better architectures such as \cite{nauman_bigger_2024},  \cite{lee2025simba}, \cite{otto2021differentiable} might be transferable to our algorithm and the rich literature on architectural improvements in off-policy RL can be tested in on-policy value learning.

\section*{Acknowledgments}

CV acknowledges funding through the Ontario Graduate Scholarship.
AMF acknowledges the support of NSERC through the Discovery Grant program [2021-03701], Polytechnique Montréal’s PIED program, and IVADO IAR$^3$ grant.
Resources used in preparing this research were provided, in part, by the Province of Ontario, the Government of Canada through CIFAR, and companies sponsoring the Vector Institute.
EE and MH’s research was partially supported by the DARPA Triage Challenge under award HR00112420305. Any opinions, findings, and conclusion or recommendations expressed in this material are those of the authors and do not necessarily reflect the view of DARPA or the US government.

\bibliography{grand_bib}

@inproceedings{li2018policy,
  title={Policy optimization with second-order advantage information},
  author={Li, Jiajin and Wang, Baoxiang and Zhang, Shengyu},
  booktitle={Proceedings of the International Joint Conference on Artificial Intelligence},
  year={2018}
}

@article{makoviychuk2021isaac,
  title={Isaac gym: High performance gpu-based physics simulation for robot learning},
  author={Makoviychuk, Viktor and Wawrzyniak, Lukasz and Guo, Yunrong and Lu, Michelle and Storey, Kier and Macklin, Miles and Hoeller, David and Rudin, Nikita and Allshire, Arthur and Handa, Ankur and others},
  journal={arXiv preprint arXiv:2108.10470},
  year={2021}
}

@article{radosavovic2024real,
  title={Real-world humanoid locomotion with reinforcement learning},
  author={Radosavovic, Ilija and Xiao, Tete and Zhang, Bike and Darrell, Trevor and Malik, Jitendra and Sreenath, Koushil},
  journal={Science Robotics},
  volume={9},
  number={89},
  pages={eadi9579},
  year={2024},
  publisher={American Association for the Advancement of Science}
}

@article{kaufmann2023champion,
  title={Champion-level drone racing using deep reinforcement learning},
  author={Kaufmann, Elia and Bauersfeld, Leonard and Loquercio, Antonio and M{\"u}ller, Matthias and Koltun, Vladlen and Scaramuzza, Davide},
  journal={Nature},
  volume={620},
  number={7976},
  pages={982--987},
  year={2023},
  publisher={Nature Publishing Group UK London}
}

@inproceedings{rudin2022learning,
  title={Learning to walk in minutes using massively parallel deep reinforcement learning},
  author={Rudin, Nikita and Hoeller, David and Reist, Philipp and Hutter, Marco},
  booktitle={Conference on robot learning},
  pages={91--100},
  year={2022},
  organization={PMLR}
}

@inproceedings{gao2023scaling,
  title={Scaling laws for reward model overoptimization},
  author={Gao, Leo and Schulman, John and Hilton, Jacob},
  booktitle={Proceedings of the International Conference on Machine Learning},
  year={2023},
}

@article{touvron2023llama,
  title={Llama 2: Open foundation and fine-tuned chat models},
  author={Touvron, Hugo and Martin, Louis and Stone, Kevin and Albert, Peter and Almahairi, Amjad and Babaei, Yasmine and Bashlykov, Nikolay and Batra, Soumya and Bhargava, Prajjwal and Bhosale, Shruti and others},
  journal={arXiv preprint arXiv:2307.09288},
  year={2023}
}

@article{liu2024deepseek,
  title={Deepseek-v3 technical report},
  author={Liu, Aixin and Feng, Bei and Xue, Bing and Wang, Bingxuan and Wu, Bochao and Lu, Chengda and Zhao, Chenggang and Deng, Chengqi and Zhang, Chenyu and Ruan, Chong and others},
  journal={arXiv preprint arXiv:2412.19437},
  year={2024}
}

@article{ouyang2022training,
  title={Training language models to follow instructions with human feedback},
  author={Ouyang, Long and Wu, Jeffrey and Jiang, Xu and Almeida, Diogo and Wainwright, Carroll and Mishkin, Pamela and Zhang, Chong and Agarwal, Sandhini and Slama, Katarina and Ray, Alex and others},
  journal={Advances in neural information processing systems},
  volume={35},
  year={2022}
}

@article{rahn2023policy,
  title={Policy optimization in a noisy neighborhood: On return landscapes in continuous control},
  author={Rahn, Nate and D'Oro, Pierluca and Wiltzer, Harley and Bacon, Pierre-Luc and Bellemare, Marc},
  journal={Advances in Neural Information Processing Systems},
  volume={36},
  pages={30618--30640},
  year={2023}
}

@inproceedings{abbas_loss_2023,
	title = {Loss of Plasticity in Continual Deep Reinforcement Learning},
	booktitle = {Proceedings of the Conference on Lifelong Learning Agents},
	author = {Abbas, Zaheer and Zhao, Rosie and Modayil, Joseph and White, Adam and Machado, Marlos C.},
	year = {2023},
}

@inproceedings{agarwal_deep_2021,
	title = {Deep Reinforcement Learning at the Edge of the Statistical Precipice},
	booktitle = {Advances in Neural Information Processing Systems},
	author = {Agarwal, Rishabh and Schwarzer, Max and Castro, Pablo Samuel and Courville, Aaron and Bellemare, Marc G.},
	year = {2021},
}

@inproceedings{ba_layer_2016,
	title = {Layer Normalization},
	volume = {abs/1607.06450},
	booktitle = {{ArXiv}},
	author = {Ba, Jimmy Lei and Kiros, Jamie Ryan and Hinton, Geoffrey E.},
	year = {2016},
}

@incollection{baird_residual_1995,
	title = {Residual algorithms: Reinforcement learning with function approximation},
	booktitle = {Machine Learning},
	publisher = {Springer},
	author = {Baird, Leemon},
	year = {1995},
}

@inproceedings{ball_efficient_2023,
	title = {Efficient online reinforcement learning with offline data},
	booktitle = {Proceedings of the International Conference on Machine Learning},
	author = {Ball, Philip J. and Smith, Laura and Kostrikov, Ilya and Levine, Sergey},
	year = {2023},
}

@inproceedings{bellemare_distributional_2017,
	title = {A Distributional Perspective on Reinforcement Learning},
	booktitle = {Proceedings of the International Conference on Machine Learning},
	author = {Bellemare, Marc G. and Dabney, Will and Munos, Rémi},
	year = {2017},
}

@inproceedings{chan_measuring_2020,
	title = {Measuring the Reliability of Reinforcement Learning Algorithms},
	booktitle = {Proceedings of the International Conference on Learning Representations},
	author = {Chan, Stephanie and Fishman, Sam and Canny, John and Korattikara, Anoop and Guadarrama, Sergio},
	year = {2020},
}

@inproceedings{lillicrap_continuous_2016,
	title = {Continuous control with deep reinforcement learning},
	booktitle = {Proceedings of the International Conference on Learning Representations},
	author = {Lillicrap, Timothy P and Hunt, Jonathan J and Pritzel, Alexander and Heess, Nicolas and Erez, Tom and Tassa, Yuval and Silver, David and Wierstra, Daan},
	year = {2016},
}

@inproceedings{doro_sample-efficient_2023,
	title = {Sample-Efficient Reinforcement Learning by Breaking the Replay Ratio Barrier},
	booktitle = {Proceedings of the International Conference on Learning Representations},
	author = {D'Oro, Pierluca and Schwarzer, Max and Nikishin, Evgenii and Bacon, Pierre-Luc and Bellemare, Marc G. and Courville, Aaron},
	year = {2023},
}

@inproceedings{farebrother_stop_2024,
	title = {Stop Regressing: Training Value Functions via Classification for Scalable Deep {RL}},
	booktitle = {Proceedings of the International Conference on Machine Learning},
	author = {Farebrother, Jesse and Orbay, Jordi and Vuong, Quan and Taiga, Adrien Ali and Chebotar, Yevgen and Xiao, Ted and Irpan, Alex and Levine, Sergey and Castro, Pablo Samuel and Faust, Aleksandra and Kumar, Aviral and Agarwal, Rishabh},
	year = {2024},
}

@inproceedings{fedus_revisiting_2020,
	title = {Revisiting fundamentals of experience replay},
	booktitle = {Proceedings of the International Conference on Machine Learning},
	author = {Fedus, William and Ramachandran, Prajit and Agarwal, Rishabh and Bengio, Yoshua and Larochelle, Hugo and Rowland, Mark and Dabney, Will},
	year = {2020},
}

@inproceedings{fujimoto_addressing_2018,
	title = {Addressing Function Approximation Error in Actor-Critic Methods},
	booktitle = {Proceedings of the International Conference on Machine Learning},
	author = {Fujimoto, Scott and van Hoof, Herke and Meger, David},
	year = {2018},
}

@inproceedings{hafner_mastering_2021,
	title = {Mastering Atari with Discrete World Models},
	booktitle = {Proceedings of the International Conference on Learning Representations},
	author = {Hafner, Danijar and Lillicrap, Timothy P. and Norouzi, Mohammad and Ba, Jimmy},
	year = {2021},
}

@inproceedings{hansen_td-mpc2_2024,
	title = {{TD}-{MPC}2: Scalable, Robust World Models for Continuous Control},
	booktitle = {Proceedings of the International Conference on Learning Representations},
	author = {Hansen, Nicklas and Su, Hao and Wang, Xiaolong},
	year = {2024},
}

@inproceedings{van_hasselt_double_2010,
	title = {Double Q-learning},
	booktitle = {Advances in Neural Information Processing Systems},
	author = {Van Hasselt, Hado},
	year = {2010},
}

@inproceedings{hussing_dissecting_2024,
	title = {Dissecting Deep {RL} with High Update Ratios: Combatting Value Divergence},
	booktitle = {Reinforcement Learning Conference},
	author = {Hussing, Marcel and Voelcker, Claas and Gilitschenski, Igor and Farahmand, Amir-massoud and Eaton, Eric},
	year = {2024},
}

@inproceedings{jaderberg_reinforcement_2017,
	title = {Reinforcement Learning with Unsupervised Auxiliary Tasks},
	booktitle = {Proceedings of the International Conference on Learning Representations},
	author = {Jaderberg, Max and Mnih, Volodymyr and Czarnecki, Wojciech Marian and Schaul, Tom and Leibo, Joel Z. and Silver, David and Kavukcuoglu, Koray},
	year = {2017},
}

@inproceedings{kumar_implicit_2021,
	title = {Implicit Under-Parameterization Inhibits Data-Efficient Deep Reinforcement Learning},
	booktitle = {Proceedings of the International Conference on Learning Representations},
	author = {Kumar, Aviral and Agarwal, Rishabh and Ghosh, Dibya and Levine, Sergey},
	year = {2021},
}

@inproceedings{lyle_understanding_2023,
	title = {Understanding Plasticity in Neural Networks},
	booktitle = {Proceedings of the International Conference on Machine Learning},
	author = {Lyle, Clare and Zheng, Zeyu and Nikishin, Evgenii and Avila Pires, Bernardo and Pascanu, Razvan and Dabney, Will},
	year = {2023},
}

@inproceedings{lyle_disentangling_2024,
	title = {Disentangling the Causes of Plasticity Loss in Neural Networks},
	author = {Lyle, Clare and Zheng, Zeyu and Khetarpal, Khimya and Hasselt, Hado Van and Pascanu, Razvan and Martens, James and Dabney, Will},
	year = {2024},
	booktitle = {Proceedings of the Conference on Lifelong Learning Agents},
}

@inproceedings{moskovitz_tactical_2021,
	title = {Tactical optimism and pessimism for deep reinforcement learning},
	booktitle = {Advances in Neural Information Processing Systems},
	author = {Moskovitz, Ted and Parker-Holder, Jack and Pacchiano, Aldo and Arbel, Michael and Jordan, Michael},
	year = {2021},
}

@inproceedings{nauman_bigger_2024,
	title = {Bigger, Regularized, Optimistic: scaling for compute and sample-efficient continuous control},
	booktitle = {Advances in Neural Information Processing Systems},
	author = {Nauman, Michal and Ostaszewski, Mateusz and Jankowski, Krzysztof and Miłoś, Piotr and Cygan, Marek},
	year = {2024},
}

@inproceedings{nauman_overestimation_2024,
	title = {Overestimation, Overfitting, and Plasticity in Actor-Critic: the Bitter Lesson of Reinforcement Learning},
	booktitle = {Proceedings of the International Conference on Machine Learning},
	author = {Nauman, Michal and Bortkiewicz, Michał and Miłoś, Piotr and Trzcinski, Tomasz and Ostaszewski, Mateusz and Cygan, Marek},
	year = {2024},
}

@inproceedings{
nauman2025bigger,
title={Bigger, Regularized, Categorical: High-Capacity Value Functions are Efficient Multi-Task Learners},
author={Michal Nauman and Marek Cygan and Carmelo Sferrazza and Aviral Kumar and Pieter Abbeel},
booktitle={The Thirty-ninth Annual Conference on Neural Information Processing Systems},
year={2025},
url={https://openreview.net/forum?id=zhOUfuOIzA}
}

@inproceedings{ni_bridging_2024,
	title = {Bridging State and History Representations: Understanding Self-Predictive {RL}},
	booktitle = {Proceedings of the International Conference on Learning Representations},
	author = {Ni, Tianwei and Eysenbach, Benjamin and Seyedsalehi, Erfan and Ma, Michel and Gehring, Clement and Mahajan, Aditya and Bacon, Pierre-Luc},
	year = {2024},
}

@inproceedings{nikishin_primacy_2022,
	title = {The Primacy Bias in Deep Reinforcement Learning},
	booktitle = {Proceedings of the International Conference on Machine Learning},
	author = {Nikishin, Evgenii and Schwarzer, Max and D'Oro, Pierluca and Bacon, Pierre-Luc and Courville, Aaron},
	year = {2022},
}

@inproceedings{peters_natural_2008,
	title = {Natural actor-critic},
	volume = {71},
	booktitle = {Neurocomputing},
	publisher = {Elsevier},
	author = {Peters, Jan and Schaal, Stefan},
	year = {2008},
}

@inproceedings{paszke_pytorch_2019,
	title = {{PyTorch}: An Imperative Style, High-Performance Deep Learning Library},
	booktitle = {Advances in Neural Information Processing Systems},
	author = {Paszke, Adam and Gross, Sam and Massa, Francisco and Lerer, Adam and Bradbury, James and Chanan, Gregory and Killeen, Trevor and Lin, Zeming and Gimelshein, Natalia and Antiga, Luca and Desmaison, Alban and Kopf, Andreas and Yang, Edward and {DeVito}, Zachary and Raison, Martin and Tejani, Alykhan and Chilamkurthy, Sasank and Steiner, Benoit and Fang, Lu and Bai, Junjie and Chintala, Soumith},
	year = {2019},
}

@inproceedings{haarnoja_soft_2018,
	title = {Soft Actor-Critic: Off-policy maximum entropy deep reinforcement learning with a stochastic actor},
	booktitle = {Proceedings of the International Conference on Machine Learning},
	author = {Haarnoja, Tuomas and Zhou, Aurick and Abbeel, Pieter and Levine, Sergey},
	year = {2018},
}

@inproceedings{schwarzer_data-efficient_2021,
	title = {Data-Efficient Reinforcement Learning with Self-Predictive Representations},
	booktitle = {Proceedings of the International Conference on Learning Representations},
	author = {Schwarzer, Max and Anand, Ankesh and Goel, Rishab and Hjelm, R. Devon and Courville, Aaron and Bachman, Philip},
	year = {2021},
}

@inproceedings{schwarzer_bigger_2023,
	title = {Bigger, Better, Faster: Human-level Atari with human-level efficiency},
	booktitle = {Proceedings of the International Conference on Machine Learning},
	author = {Schwarzer, Max and Obando Ceron, Johan Samir and Courville, Aaron and Bellemare, Marc G and Agarwal, Rishabh and Castro, Pablo Samuel},
	year = {2023},
}

@inproceedings{silver_deterministic_2014,
	title = {Deterministic policy gradient algorithms},
	booktitle = {Proceedings of the International Conference on Machine Learning},
	author = {Silver, David and Lever, Guy and Heess, Nicolas and Degris, Thomas and Wierstra, Daan and Riedmiller, Martin},
	year = {2014},
}

@inproceedings{sutton_learning_1988,
	title = {Learning to predict by the methods of temporal differences},
	volume = {3},
	booktitle = {Machine learning},
	publisher = {Springer},
	author = {Sutton, Richard S},
	year = {1988},
}

@inproceedings{sutton_emphatic_2016,
	title = {An emphatic approach to the problem of off-policy temporal-difference learning},
	volume = {17},
	booktitle = {Journal of Machine Learning Research},
	publisher = {{MIT} Press},
	author = {Sutton, Richard S and Mahmood, A Rupam and White, Martha},
	year = {2016},
}

@book{sutton_reinforcement_2018,
	location = {Cambridge, {MA}, {USA}},
	edition = {2nd},
	title = {Reinforcement Learning: An Introduction},
	publisher = {A Bradford Book},
	author = {Sutton, Richard S. and Barto, Andrew G.},
	year = {2018},
}

@inproceedings{tang_understanding_2023,
	title = {Understanding Self-Predictive Learning for Reinforcement Learning},
	booktitle = {Proceedings of the International Conference on Machine Learning},
	author = {Tang, Yunhao and Guo, Zhaohan Daniel and Richemond, Pierre Harvey and Pires, Bernardo Ávila and Chandak, Yash and Munos, Rémi and Rowland, Mark and Azar, Mohammad Gheshlaghi and Lan, Charline Le and Lyle, Clare and {others}},
	year = {2023},
}

@inproceedings{thrun_issues_1993,
	title = {Issues in Using Function Approximation for Reinforcement Learning},
	booktitle = {Connectionist Models Summer School},
	author = {Thrun, Sebastian and Schwartz, Anton},
	year = {1993},
}

@inproceedings{voelcker_when_2024,
	title = {When does self-prediction help? Understanding Auxiliary Tasks in Reinforcement Learning},
	booktitle = {Reinforcement Learning Conference},
	author = {Voelcker, Claas and Kastner, Tyler and Gilitschenski, Igor and Farahmand, Amir-massoud},
	year = {2024},
}

@inproceedings{voelcker_mad-td_2025,
	title = {{MAD}-{TD}: Model-Augmented Data stabilizes High Update Ratio {RL}},
	booktitle = {Proceedings of the International Conference on Learning Representations},
	author = {Voelcker, Claas and Hussing, Marcel and Eaton, Eric and Farahmand, Amir-massoud and Gilitschenski, Igor},
	year = {2025},
}

@inproceedings{voelcker_can_2024,
	title = {Can we hop in general? A discussion of benchmark selection and design using the Hopper environment},
	booktitle = {Finding the Frame: An {RLC} Workshop for Examining Conceptual Frameworks},
	author = {Voelcker, Claas and Hussing, Marcel and Eaton, Eric},
	year = {2024},
}

@inproceedings{suh_differentiable_2022,
	title = {Do differentiable simulators give better policy gradients?},
	booktitle = {Proceedings of the International Conference on Machine Learning},
	author = {Suh, Hyung Ju and Simchowitz, Max and Zhang, Kaiqing and Tedrake, Russ},
	year = {2022},
}

@article{levine_reinforcement_2018,
	title = {Reinforcement learning and control as probabilistic inference: Tutorial and review},
	journal = {{arXiv} preprint {arXiv}:1805.00909},
	author = {Levine, Sergey},
	year = {2018},
}

@misc{zakka_mujoco_2025,
	title = {{MuJoCo} Playground: An open-source framework for {GPU}-accelerated robot learning and sim-to-real transfer.},
	url = {https://github.com/google-deepmind/mujoco_playground},
	shorttitle = {{MuJoCo} Playground},
	author = {Zakka, Kevin and Tabanpour, Baruch and Liao, Qiayuan and Haiderbhai, Mustafa and Holt, Samuel and Luo, Jing Yuan and Allshire, Arthur and Frey, Erik and Sreenath, Koushil and Kahrs, Lueder A. and Sferrazza, Carlo and Tassa, Yuval and Abbeel, Pieter},
	year = {2025},
}

@article{seo2025fasttd3,
  title={FastTD3: Simple, Fast, and Capable Reinforcement Learning for Humanoid Control},
  author={Seo, Younggyo and Sferrazza, Carmelo and Geng, Haoran and Nauman, Michal and Yin, Zhao-Heng and Abbeel, Pieter},
  journal={arXiv preprint arXiv:2505.22642},
  year={2025}
}

@inproceedings{li_parallel_2023,
	title = {Parallel Q-Learning: Scaling Off-policy Reinforcement Learning under Massively Parallel Simulation},
	booktitle = {Proceedings of the International Conference on Machine Learning},
	author = {Li, Zechu and Chen, Tao and Hong, Zhang-Wei and Ajay, Anurag and Agrawal, Pulkit},
	year = {2023},
}

@inproceedings{gallici_simplifying_2024,
	title = {Simplifying Deep Temporal Difference Learning},
	booktitle = {Proceedings of the International Conference on Learning Representations},
	author = {Gallici, Matteo and Fellows, Mattie and Ellis, Benjamin and Pou, Bartomeu and Masmitja, Ivan and Foerster, Jakob Nicolaus and Martin, Mario},
	year = {2024},
	langid = {english},
}

@inproceedings{peters_relative_2010,
	location = {Atlanta, Georgia},
	title = {Relative entropy policy search},
	booktitle = {Proceedings of the {AAAI} Conference on Artificial Intelligence},
	author = {Peters, Jan and Mülling, Katharina and Altün, Yasemin},
	year = {2010},
}

@inproceedings{song_v-mpo_2019,
	title = "{V-{MPO}: On-Policy Maximum a Posteriori Policy Optimization for Discrete and Continuous Control}",
	booktitle = {Proceedings of the International conference on Learning Representations},
	author = {Song, H. Francis and Abdolmaleki, Abbas and Springenberg, Jost Tobias and Clark, Aidan and Soyer, Hubert and Rae, Jack W. and Noury, Seb and Ahuja, Arun and Liu, Siqi and Tirumala, Dhruva and Heess, Nicolas and Belov, Dan and Riedmiller, Martin and Botvinick, Matthew M.},
	year = {2019},
}

@inproceedings{abdolmaleki_maximum_2018,
	title = {Maximum a Posteriori Policy Optimisation},
	booktitle = {Proceedings of the International Conference on Learning Representations},
	author = {Abdolmaleki, Abbas and Springenberg, Jost Tobias and Tassa, Yuval and Munos, Remi and Heess, Nicolas and Riedmiller, Martin},
	year = {2018},
}

@misc{schulman_proximal_2017,
	title = {Proximal Policy Optimization Algorithms},
	doi = {10.48550/arXiv.1707.06347},
	number = {{arXiv}:1707.06347},
	publisher = {{arXiv}},
	author = {Schulman, John and Wolski, Filip and Dhariwal, Prafulla and Radford, Alec and Klimov, Oleg},
	year = {2017},
	eprinttype = {arxiv},
	eprint = {1707.06347 [cs]},
}

@inproceedings{schulman_trust_2015,
	title = {Trust Region Policy Optimization},
	eventtitle = {Proceedings of the International Conference on Machine Learning},
	booktitle = {Proceedings of the International Conference on Machine Learning},
	publisher = {{PMLR}},
	author = {Schulman, John and Levine, Sergey and Abbeel, Pieter and Jordan, Michael and Moritz, Philipp},
	year = {2015},
}

@article{taomaniskill3,
  title={ManiSkill3: GPU Parallelized Robotics Simulation and Rendering for Generalizable Embodied AI},
  author={Stone Tao and Fanbo Xiang and Arth Shukla and Yuzhe Qin and Xander Hinrichsen and Xiaodi Yuan and Chen Bao and Xinsong Lin and Yulin Liu and Tse-kai Chan and Yuan Gao and Xuanlin Li and Tongzhou Mu and Nan Xiao and Arnav Gurha and Viswesh Nagaswamy Rajesh and Yong Woo Choi and Yen-Ru Chen and Zhiao Huang and Roberto Calandra and Rui Chen and Shan Luo and Hao Su},
  journal = {Robotics: Science and Systems},
  year={2025},
}

@inproceedings{daley_2019_reconciling,
author = {Daley, Brett and Amato, Christopher},
 booktitle = {Advances in Neural Information Processing Systems},
 title = {Reconciling $\lambda$-Returns with Experience Replay},
 year = {2019}
}

@article{haarnoja_soft_2019,
	title = {Soft Actor-Critic Algorithms and Applications},
    journal={arXiv preprint arXiv:1812.05905},
	publisher = {{arXiv}},
	author = {Haarnoja, Tuomas and Zhou, Aurick and Hartikainen, Kristian and Tucker, George and Ha, Sehoon and Tan, Jie and Kumar, Vikash and Zhu, Henry and Gupta, Abhishek and Abbeel, Pieter and Levine, Sergey},
	year = {2019},
}

@inproceedings{fujimoto_towards_2024,
	title = {Towards General-Purpose Model-Free Reinforcement Learning},
	booktitle = {Proceedings of the International Conference on Learning Representations},
	author = {Fujimoto, Scott and D'Oro, Pierluca and Zhang, Amy and Tian, Yuandong and Rabbat, Michael},
	year = {2024},
}

@article{mohamed2020monte,
  title={Monte carlo gradient estimation in machine learning},
  author={Mohamed, Shakir and Rosca, Mihaela and Figurnov, Michael and Mnih, Andriy},
  journal={Journal of Machine Learning Research},
  volume={21},
  number={132},
  pages={1--62},
  year={2020}
}

@inproceedings{
lee2025simba,
title={SimBa: Simplicity Bias for Scaling Up Parameters in Deep Reinforcement Learning},
author={Hojoon Lee and Dongyoon Hwang and Donghu Kim and Hyunseung Kim and Jun Jet Tai and Kaushik Subramanian and Peter R. Wurman and Jaegul Choo and Peter Stone and Takuma Seno},
booktitle={Proceedings of the International Conference on Learning Representations},
year={2025},
}

@inproceedings{
lee2025hyperspherical,
title={Hyperspherical Normalization for Scalable Deep Reinforcement Learning},
author={Hojoon Lee and Youngdo Lee and Takuma Seno and Donghu Kim and Peter Stone and Jaegul Choo},
booktitle={Proceedings of the International Conference on Machine Learning},
year={2025},
}

@inproceedings{ilyas202acloser,
title={A Closer Look at Deep Policy Gradients},
author={Andrew Ilyas and Logan Engstrom and Shibani Santurkar and Dimitris Tsipras and Firdaus Janoos and Larry Rudolph and Aleksander Madry},
booktitle={Proceedings of the International Conference on Learning Representations},
year={2020},
}

@article{greensmith2004variance,
author = {Greensmith, Evan and Bartlett, Peter L. and Baxter, Jonathan},
title = {Variance Reduction Techniques for Gradient Estimates in Reinforcement Learning},
year = {2004},
volume = {5},
journal = {Journal of Machine Learning Research},
}

@book{puterman1994markov,
author = {Puterman, Martin L.},
title = {Markov Decision Processes: Discrete Stochastic Dynamic Programming},
year = {1994},
isbn = {0471619779},
publisher = {John Wiley \& Sons, Inc.},
address = {USA},
edition = {1st},
}

@InProceedings{imani2018improving,
  title = 	 {Improving Regression Performance with Distributional Losses},
  author =       {Imani, Ehsan and White, Martha},
  booktitle = 	 {Proceedings of the International Conference on Machine Learning},
  year = 	 {2018},
}

@inproceedings{yue2023understanding,
    title={Understanding, Predicting and Better Resolving Q-Value Divergence in Offline-{RL}},
    author={Yang Yue and Rui Lu and Bingyi Kang and Shiji Song and Gao Huang},
    booktitle={Advances in Neural Information Processing Systems},
    year={2023},
}

@inproceedings{ziebart2008maximum,
    author = {Ziebart, Brian D. and Maas, Andrew and Bagnell, J. Andrew and Dey, Anind K.},
    title = {Maximum entropy inverse reinforcement learning},
    year = {2008},
    booktitle = {Proceedings of the National Conference on Artificial Intelligence},
}

@inproceedings{wang2020truly,
  title={Truly proximal policy optimization},
  author={Wang, Yuhui and He, Hao and Tan, Xiaoyang},
  booktitle={Uncertainty in Artificial Intelligence},
  year={2020},
}

@article{kakade2001natural,
  title={A natural policy gradient},
  author={Kakade, Sham M},
  journal={Advances in neural information processing systems},
  year={2001}
}

@inproceedings{nauman2025decoupled,
  title={Decoupled policy actor-critic: Bridging pessimism and risk awareness in reinforcement learning},
  author={Nauman, Michal and Cygan, Marek},
  booktitle={Proceedings of the AAAI Conference on Artificial Intelligence},
  year={2025}
}

@inproceedings{xie2025simple,
    title={Simple Policy Optimization},
    author={Zhengpeng Xie and Qiang Zhang and Fan Yang and Marco Hutter and Renjing Xu},
    booktitle={Proceedings of the International Conference on Machine Learning},
    year={2025},
}

@article{papini2024policy,
    title={Policy Gradient with Active Importance Sampling},
    author={Papini, Matteo and Manganini, Giorgio and Metelli, Alberto Maria and Restelli, Marcello},
    journal={Reinforcement Learning Journal},
    volume={2},
    pages={645--675},
    year={2024}
}

@inproceedings{grudzien2022mirror,
  title={Mirror learning: A unifying framework of policy optimisation},
  author={Grudzien, Jakub and De Witt, Christian A Schroeder and Foerster, Jakob},
  booktitle={Proceedings of the International Conference on Machine Learning},
  year={2022},
}

@article{lu2022discovered,
  title={Discovered policy optimisation},
  author={Lu, Chris and Kuba, Jakub and Letcher, Alistair and Metz, Luke and Schroeder de Witt, Christian and Foerster, Jakob},
  journal={Advances in Neural Information Processing Systems},
  year={2022}
}

@inproceedings{tomar2022mirror,
  title={Mirror Descent Policy Optimization},
  author={Tomar, Manan and Shani, Lior and Efroni, Yonathan and Ghavamzadeh, Mohammad},
  year={2022},
  booktitle={Proceedings of the International Conference on Learning Representations}
}

@inproceedings{georgiev2024adaptive,
  title={Adaptive Horizon Actor-Critic for Policy Learningin Contact-Rich Differentiable Simulation}, 
  author={Georgiev, Ignat and Srinivasan, Krishnan and Xu, Jie and Heiden, Eric and Garg, Animesh},
  booktitle={Proceedings of the International Conference on Machine Learning},
  year={2024},
  organization={PMLR},
}

@inproceedings{
xu2022accelerated,
title={Accelerated Policy Learning with Parallel Differentiable Simulation},
author={Jie Xu and Miles Macklin and Viktor Makoviychuk and Yashraj Narang and Animesh Garg and Fabio Ramos and Wojciech Matusik},
booktitle={Proceedings of the International Conference on Learning Representations},
year={2022},
}

@InProceedings{mora2021pods,
  title = 	 {PODS: Policy Optimization via Differentiable Simulation},
  author =       {Mora, Miguel Angel Zamora and Peychev, Momchil and Ha, Sehoon and Vechev, Martin and Coros, Stelian},
  booktitle = 	 {Proceedings of the International Conference on Machine Learning},
  year = 	 {2021},
}

@inproceedings{son2023gradient,
    title={Gradient Informed Proximal Policy Optimization},
    author={Sanghyun Son and Laura Yu Zheng and Ryan Sullivan and Yi-Ling Qiao and Ming Lin},
    booktitle={Advances in Neural Information Processing Systems},
    year={2023},
}

@InProceedings{cobbe2021phasic,
  title = 	 {Phasic Policy Gradient},
  author =       {Cobbe, Karl W and Hilton, Jacob and Klimov, Oleg and Schulman, John},
  booktitle = 	 {Proceedings of the 38th International Conference on Machine Learning},
  year = 	 {2021},
}

@inproceedings{aitchison2022dna,
    title={{DNA}: Proximal Policy Optimization with a Dual Network Architecture},
    author={Matthew Aitchison and Penny Sweetser},
    booktitle={Advances in Neural Information Processing Systems},
    year={2022},
}

@inproceedings{moalla2024no,
    title={No Representation, No Trust: Connecting Representation, Collapse, and Trust Issues in {PPO}},
    author={Skander Moalla and Andrea Miele and Daniil Pyatko and Razvan Pascanu and Caglar Gulcehre},
    booktitle={Advances in Neural Information Processing Systems},
    year={2024},
}

@article{abdolmaleki2015model,
  title={Model-based relative entropy stochastic search},
  author={Abdolmaleki, Abbas and Lioutikov, Rudolf and Peters, Jan R and Lau, Nuno and Pualo Reis, Luis and Neumann, Gerhard},
  journal={Advances in Neural Information Processing Systems},
  year={2015}
}

@misc{jax2018github,
  author = {James Bradbury and Roy Frostig and Peter Hawkins and Matthew James Johnson and Chris Leary and Dougal Maclaurin and George Necula and Adam Paszke and Jake Vander{P}las and Skye Wanderman-{M}ilne and Qiao Zhang},
  title = {{JAX}: composable transformations of {P}ython+{N}um{P}y programs},
  url = {http://github.com/jax-ml/jax},
  version = {0.3.13},
  year = {2018},
}

@inproceedings{
otto2021differentiable,
title={Differentiable Trust Region Layers for Deep Reinforcement Learning},
author={Fabian Otto and Philipp Becker and Vien Anh Ngo and Hanna Carolin Maria Ziesche and Gerhard Neumann},
booktitle={Proceedings of the International Conference on Learning Representations},
year={2021},
}

@inproceedings{kakade2002approximately,
author = {Kakade, Sham and Langford, John},
title = {Approximately Optimal Approximate Reinforcement Learning},
year = {2002},
booktitle = {Proceedings of the International Conference on Machine Learning},
}

@InProceedings{akrour2019projections,
  title = 	 {Projections for Approximate Policy Iteration Algorithms},
  author =       {Akrour, Riad and Pajarinen, Joni and Peters, Jan and Neumann, Gerhard},
  booktitle = 	 {Proceedings of the International Conference on Machine Learning},
  year = 	 {2019},
}

@article{pajarinen2019compatible,
  title={Compatible natural gradient policy search},
  author={Pajarinen, Joni and Thai, Hong Linh and Akrour, Riad and Peters, Jan and Neumann, Gerhard},
  journal={Machine Learning},
  volume={108},
  number={8},
  year={2019},
  publisher={Springer}
}

@inproceedings{nota202is,
author = {Nota, Chris and Thomas, Philip S.},
title = {Is the Policy Gradient a Gradient?},
year = {2020},
booktitle = {Proceedings of the International Conference on Autonomous Agents and MultiAgent Systems},
}

@inproceedings{jordan2020evaluating,
  title={Evaluating the performance of reinforcement learning algorithms},
  author={Jordan, Scott and Chandak, Yash and Cohen, Daniel and Zhang, Mengxue and Thomas, Philip},
  booktitle={Proceedings of the International Conference on Machine Learning},
  year={2020},
}

@inproceedings{chan2020measuring,
  title={Measuring the reliability of reinforcement learning algorithms},
  author={Chan, Stephanie CY and Fishman, Samuel and Canny, John and Korattikara, Anoop and Guadarrama, Sergio},
  booktitle={Proceedings of the International Conference on Learning Representations},
  year={2020}
}

@misc{
rahman2023robust,
title={Robust Policy Optimization in Deep Reinforcement Learning},
author={Md Masudur Rahman and Yexiang Xue},
year={2023},
}

@inproceedings{
celik2025dime,
title={{DIME}: Diffusion-Based Maximum Entropy Reinforcement Learning},
author={Onur Celik and Zechu Li and Denis Blessing and Ge Li and Daniel Palenicek and Jan Peters and Georgia Chalvatzaki and Gerhard Neumann},
booktitle={Proceedings of the International Conference on Machine Learning},
year={2025},
}

@inproceedings{
ma2025efficient,
title={Efficient Online Reinforcement Learning for Diffusion Policy},
author={Haitong Ma and Tianyi Chen and Kai Wang and Na Li and Bo Dai},
booktitle={Proceedings of the International Conference on Machine Learning},
year={2025},
}

@article{chi2024diffusionpolicy,
	author = {Cheng Chi and Zhenjia Xu and Siyuan Feng and Eric Cousineau and Yilun Du and Benjamin Burchfiel and Russ Tedrake and Shuran Song},
	title ={Diffusion Policy: Visuomotor Policy Learning via Action Diffusion},
	journal = {The International Journal of Robotics Research},
	year = {2024},
}

@article{christodoulou2019soft,
  title={Soft actor-critic for discrete action settings},
  author={Christodoulou, Petros},
  journal={arXiv preprint arXiv:1910.07207},
  year={2019}
}

@inproceedings{
maddison2017the,
title={The Concrete Distribution: A Continuous Relaxation of Discrete Random Variables},
author={Chris J. Maddison and Andriy Mnih and Yee Whye Teh},
booktitle={Proceedings of the International Conference on Learning Representations},
year={2017},
}

@inproceedings{
jang2017categorical,
title={Categorical Reparameterization with Gumbel-Softmax},
author={Eric Jang and Shixiang Gu and Ben Poole},
booktitle={Proceedings of the International Conference on Learning Representations},
year={2017},
}

@inproceedings{neumann2011variational,
author = {Neumann, Gerhard},
title = {Variational inference for policy search in changing situations},
year = {2011},
booktitle = {Proceedings of the International Conference on International Conference on Machine Learning},
}

@inproceedings{
sokota2022a,
title={A Unified Approach to Reinforcement Learning, Quantal Response Equilibria, and Two-Player Zero-Sum Games},
author={Samuel Sokota and Ryan D'Orazio and J Zico Kolter and Nicolas Loizou and Marc Lanctot and Ioannis Mitliagkas and Noam Brown and Christian Kroer},
booktitle={Deep Reinforcement Learning Workshop NeurIPS 2022},
year={2022},
}

@article{palenicek2025xqc,
  title={XQC: Well-conditioned Optimization Accelerates Deep Reinforcement Learning},
  author={Palenicek, Daniel and Vogt, Florian and Watson, Joe and Posner, Ingmar and Peters, Jan},
  journal={arXiv preprint arXiv:2509.25174},
  year={2025}
}

@article{rslrl,
  title={RSL-RL: A Learning Library for Robotics Research},
  author={Schwarke, Clemens and Mittal, Mayank and Rudin, Nikita and Hoeller, David and Hutter, Marco},
  journal={arXiv preprint arXiv:2509.10771},
  year={2025}
}

@article{mittal2025isaaclab,
  title={Isaac Lab: A GPU-Accelerated Simulation Framework for Multi-Modal Robot Learning},
  author={Mayank Mittal and Pascal Roth and James Tigue and Antoine Richard and Octi Zhang and Peter Du and Antonio Serrano-Muñoz and Xinjie Yao and René Zurbrügg and Nikita Rudin and Lukasz Wawrzyniak and Milad Rakhsha and Alain Denzler and Eric Heiden and Ales Borovicka and Ossama Ahmed and Iretiayo Akinola and Abrar Anwar and Mark T. Carlson and Ji Yuan Feng and Animesh Garg and Renato Gasoto and Lionel Gulich and Yijie Guo and M. Gussert and Alex Hansen and Mihir Kulkarni and Chenran Li and Wei Liu and Viktor Makoviychuk and Grzegorz Malczyk and Hammad Mazhar and Masoud Moghani and Adithyavairavan Murali and Michael Noseworthy and Alexander Poddubny and Nathan Ratliff and Welf Rehberg and Clemens Schwarke and Ritvik Singh and James Latham Smith and Bingjie Tang and Ruchik Thaker and Matthew Trepte and Karl Van Wyk and Fangzhou Yu and Alex Millane and Vikram Ramasamy and Remo Steiner and Sangeeta Subramanian and Clemens Volk and CY Chen and Neel Jawale and Ashwin Varghese Kuruttukulam and Michael A. Lin and Ajay Mandlekar and Karsten Patzwaldt and John Welsh and Huihua Zhao and Fatima Anes and Jean-Francois Lafleche and Nicolas Moënne-Loccoz and Soowan Park and Rob Stepinski and Dirk Van Gelder and Chris Amevor and Jan Carius and Jumyung Chang and Anka He Chen and Pablo de Heras Ciechomski and Gilles Daviet and Mohammad Mohajerani and Julia von Muralt and Viktor Reutskyy and Michael Sauter and Simon Schirm and Eric L. Shi and Pierre Terdiman and Kenny Vilella and Tobias Widmer and Gordon Yeoman and Tiffany Chen and Sergey Grizan and Cathy Li and Lotus Li and Connor Smith and Rafael Wiltz and Kostas Alexis and Yan Chang and David Chu and Linxi "Jim" Fan and Farbod Farshidian and Ankur Handa and Spencer Huang and Marco Hutter and Yashraj Narang and Soha Pouya and Shiwei Sheng and Yuke Zhu and Miles Macklin and Adam Moravanszky and Philipp Reist and Yunrong Guo and David Hoeller and Gavriel State},
  journal={arXiv preprint arXiv:2511.04831},
  year={2025},
}
\bibliographystyle{iclr2026_conference}

\appendix

\section{Extended Related Work}

\paragraph{Stabilizing On-Policy RL}
A fundamental issue with score-based approaches is their instability. Therefore, various improvements to decrease gradient variance have been considered. Some works have noted the difficulty of representation learning and have addressed this via decoupling the training of value and policy \citep{cobbe2021phasic, aitchison2022dna}. \cite{moalla2024no} note that feature learning problems can result from representation collapse, which can be mitigated using auxiliary losses. There are also efforts to reduce the variance of gradients, e.g. by finding a policy that minimizes the variance of the importance sampling factor~\citep{papini2024policy} or modifying the loss to ensure tighter total variational distance constraints~\citep{xie2025simple}.

Incorporating ground-truth gradient signal to stabilize training has also been studied, both for dynamical systems \citep{son2023gradient} and differentiable robotics simulation \citep{mora2021pods, xu2022accelerated, georgiev2024adaptive}. 
However, access to a ground-truth gradient requires custom simulators, and in contact-rich tasks, approximate models can provide smoother gradients \citep{suh_differentiable_2022}.

\paragraph{Trust regions and constrained policy optimization}
Other approaches have used similar KL and trust region constraint as \algoname.
\cite{schulman_trust_2015} and \cite{peters_relative_2010} formulate the KL constrained policy update as a constrained optimization problem.
\cite{peters_relative_2010} shows a closed form solution to this problem, while \cite{schulman_trust_2015} uses a conjugate gradient scheme to solve the relaxed optimization problem.
\cite{schulman_proximal_2017} replaces the Lagrangian formulation with a clipping heuristic.
However, clipping can lead to wrong gradient estimates \citep{ilyas202acloser} and in some scenarios the clipping objective fails to bound the policy deviation~\citep{wang2020truly}.
\cite{akrour2019projections} propose to project the policy onto the trust-region to sidestep the difficulty associated with clipping.
We find that our approach is simpler to implement and more general, as we do not assume direct projection is possible.

\cite{otto2021differentiable} propose to replace the various trust-region enforcement methods such as line-search or clipping with differentiable trust-region layers in the policy neural network architecture.
While our method is slightly more general, as we make no assumption on the form of the policy (aside from assuming gradient propagation through the sampling process is possible), trust-region layers could easily be combined with \algoname for appropriate policy parameterizations.

\paragraph{Work on GPU-parallelized On-policy RL} With the parallelization of many benchmarks on GPUs~\citep{makoviychuk2021isaac,zakka_mujoco_2025,taomaniskill3}, massively-parallel on-policy RL has become quite popular. While these environments provide simulation testbeds, algorithms trained in such environments have shown to transfer to real-robots, allowing us to train them in minutes rather than days~\citep{rudin2022learning}.

\paragraph{Hybridizing Off-policy and On-policy RL methods}
Most closely to our work, Parallel Q Networks (PQN) \citep{gallici_simplifying_2024} was established by using standard discrete action-space off-policy techniques in the MPS setting.
While our work shares several important features with this method, we find that our additional insights on KL regularization and tuning is crucial for adapting the concept to continuous action spaces.
We also evaluate our approach on discrete action spaces (see \autoref{app:dreppo}).
While PQN performs slightly better, likely owing to tuned exploration techniques, we show that our method works robustly across both discrete \emph{and} continuous action spaces.

Other methods, such as Parallel Q-Learning \citep{li_parallel_2023} and FastTD3 \citep{seo2025fasttd3} also attempt to use deterministic policy gradient algorithms in the MPS setting, but still remain off-policy.
This has two major drawbacks compared to our work.
The methods require very large replay buffers, which can either limit the speed if data needs to be stored in regular CPU memory, or require very large and expensive GPUs.
In addition, the off-policy nature of these methods requires stabilizing techniques such as clipped double Q learning, which has been shown to prevent exploration.

\paragraph{KL-based RL}
Finally, other works also build on top of the relative entropy policy search \citep{peters_relative_2010}.
Maximum A Posteriori Policy Optimization (MPO) \citep{abdolmaleki_maximum_2018} and Variational MPO \citep{song_v-mpo_2019} both leverage SAC style maximum entropy objectives and use KL constraints to prevent policy divergence.
However, both methods use off-policy data together with importance sampling, which we forgo, do not tune the KL and entropy parameters, and crucially do not make use of the deterministic policy gradient.

Going beyond relative entropy, the KL-based constraint formulation has been generalized to include the class of mirror descent algorithms \citep{grudzien2022mirror,tomar2022mirror}.
In addition, \cite{lu2022discovered} meta-learns a constraint to automatically discover novel RL algorithms.
These advancements are largely orthogonal to our work and can be incorporated into \algoname in the future.

\paragraph{Instability in Off-policy RL} Our method furthermore adapts many design decisions from recent off-policy literature.
Among these are layer normalizations, which have been studied by \cite{nauman_overestimation_2024,hussing_dissecting_2024,nauman_bigger_2024,gallici_simplifying_2024}, auxiliary tasks \citep{jaderberg_reinforcement_2017,schwarzer_data-efficient_2021,schwarzer_bigger_2023,tang_understanding_2023,voelcker_when_2024,ni_bridging_2024}, and HL-Gauss \citep{farebrother_stop_2024}, variants of which have been used by \cite{hafner_mastering_2021,hansen_td-mpc2_2024,voelcker_mad-td_2025}.
Beyond these, there are several other works which investigate architectures for stable off-policy value learning, such as \cite{nauman_bigger_2024, lee2025simba,lee2025hyperspherical}.
A similar method to our KL regularization tuning objective has been used by \citep{nauman2025decoupled} to build an exploratory optimistic actor.
While the technique is very similar, we employ it in the context of the trust-region update, and show the importance of jointly tuning the entropy and KL parameters.
Finally, there are several papers which investigate the impact of continual learning in off-policy reinforcement learning, including issues such as out-of-distribution misgeneralization \citep{voelcker_mad-td_2025}, plasticity loss \citep{nikishin_primacy_2022,doro_sample-efficient_2023,lyle_understanding_2023,abbas_loss_2023}.
Since many of these works focus specifically on improving issues inherent in the off-policy setting, we did not evaluate all of these changes in \algoname.
However, rigorously evaluating what network architectures and stabilization methods can help to further improve the online regime is an exciting avenue for future work.

\section{Wallclock Measurement Considerations}
\label{app:wallclock}

Measuring wall-clock time has become a popular way of highlighting the practical utility of an algorithm as it allows us to quickly deploy new models and iterate on ideas.
Rigorous wall-clock time measurement is a difficult topic, as many factors impact the wall-clock time of an algorithm.

We chose to not compare the jax and torch versions head-to-head as we found significant runtime differences on different hardware, and the different compilation philosophies lead to different benefits and drawbacks.
For example, jax' full jit-compilation trades a much larger initial overhead for significantly faster execution, which can amortize itself depending on the number of timesteps taken.
This is the reason why we do not include FastTD3 in \autoref{fig:wallclock}, as only a PyTorch implementation of the algorithm exists.
FastTD3 and \algoname use similar algorithms and hyperparameters, therefore, barring complexities like those discussed below, we expect them to perform at similar speeds.

More importantly, torch's compilation libraries are built to accelerate standard supervised and generative workflows, but do not support RL primitives equally well.
As the CPU needs to load kernels during training which the GPU then executes, the CPU plays a much larger role in the speed measurements of the torch-based variant of REPPO.
Especially the tanh-squashed log probability computation and the frequent resampling from the action space cannot be offloaded into an efficient kernel without providing one manually, which we have not done.
This is likely due to the fact that torch keeps its random seed on the CPU.
This is not a concern for jax, due to the fact that all kernels are statically compiled when the program is first executed, and random seeds are handled explicitly as part of the program state.
Therefore, the CPU is under much lower load.

Instead of raw wall-clock time measurements, which can vary massively across framework and hardware, we recommend that the community treat the question of wall-clock time more carefully.
While the actual time for an experiment can be of massive importance from a practical point of view, the advantages and limitations of current frameworks can obscure exciting directions for future work. For example \algoname is highly competitive with PPO when implemented in jax, but struggles somewhat in torch due to framework specific design choices.

\section{Discrete REPPO (D-REPPO)}
\label{app:dreppo}

One of the major advantages of PPO in the zoo of RL algorithms is the fact that it can be used in both continuous and discrete action settings.
However, as we build on the DDPG/TD3/SAC line of work, the exposition of our algorithm has focused on the continuous setting alone.

Nonetheless, it is easy to adapt our approach to the discrete action setting as well.
Following the proposal of \cite{christodoulou2019soft}, we can circumvent the chained critic-actor gradient and compute the value of the current policy, the entropy, and the KL bound in closed form
\begin{align}
    \mathcal{L}^\mathrm{D-REPPO}_{\pi,\leq\mathrm{KL}}(\theta|B) &= 
    -\frac{1}{|B|} \sum_{i=1}^{|B|} \sum_{j=1}^{|A|}\pi_{\theta}(a_j|x_i)\left(Q(x_i,a_j) + e^\alpha \log \pi_{\theta}(a_j|x_i) \right)\\
    \mathcal{L}^\mathrm{D-REPPO}_{\pi,>\mathrm{KL}}(\theta|B) &= -\frac{1}{|B|} \sum_{i=1}^{|B|} e^\beta \sum_{j=1}^{|A|} \pi_{\theta'}(a_j|x_i)\log \frac{\pi_{\theta'}(a_j|x_i)}{\pi_\theta(a_j|x_i)}
\label{eq:dreppo_pi}
\end{align}
\begin{align}
    \mathcal{L}^\mathrm{D-REPPO}_\pi(\theta|B) = \begin{cases}
    \mathcal{L}^\mathrm{D-REPPO}_{\pi,\leq\mathrm{KL}}(\theta|B)  , & \text{if } \sum_{j=1}^k \log \frac{\pi_{\theta'}(a_j|x_i)}{\pi_\theta(a_j|x_i)} < \varepsilon_\mathrm{KL} \\
    \mathcal{L}^\mathrm{D-REPPO}_{\pi,>\mathrm{KL}}(\theta|B), & \text{otherwise}.
\end{cases}
\end{align}

This variant of our algorithm still directly differentiates the full Q function objective, so can still be seen as a pathwise implementation.
But computing the expectation in closed form circumvents the necessity to use a biased estimator for discrete sampling, such as the Gumbel-Softmax trick \citep{maddison2017the,jang2017categorical,fujimoto_towards_2024}.

To investigate the benefits of our approach in the discrete action setting, we compare it against PQN \citep{gallici_simplifying_2024} and PPO.
The main benefit of our approach over PQN is that it is a) a general algorithm that unifies both discrete and continuous action spaces, due to the underlying actor critic architecture, and b) that the principled entropy and KL objectives stabilize updates and encourages continuing exploration without an epsilon greedy exploration strategy.

\begin{figure}[h]
    \centering
    \includegraphics[width=\linewidth]{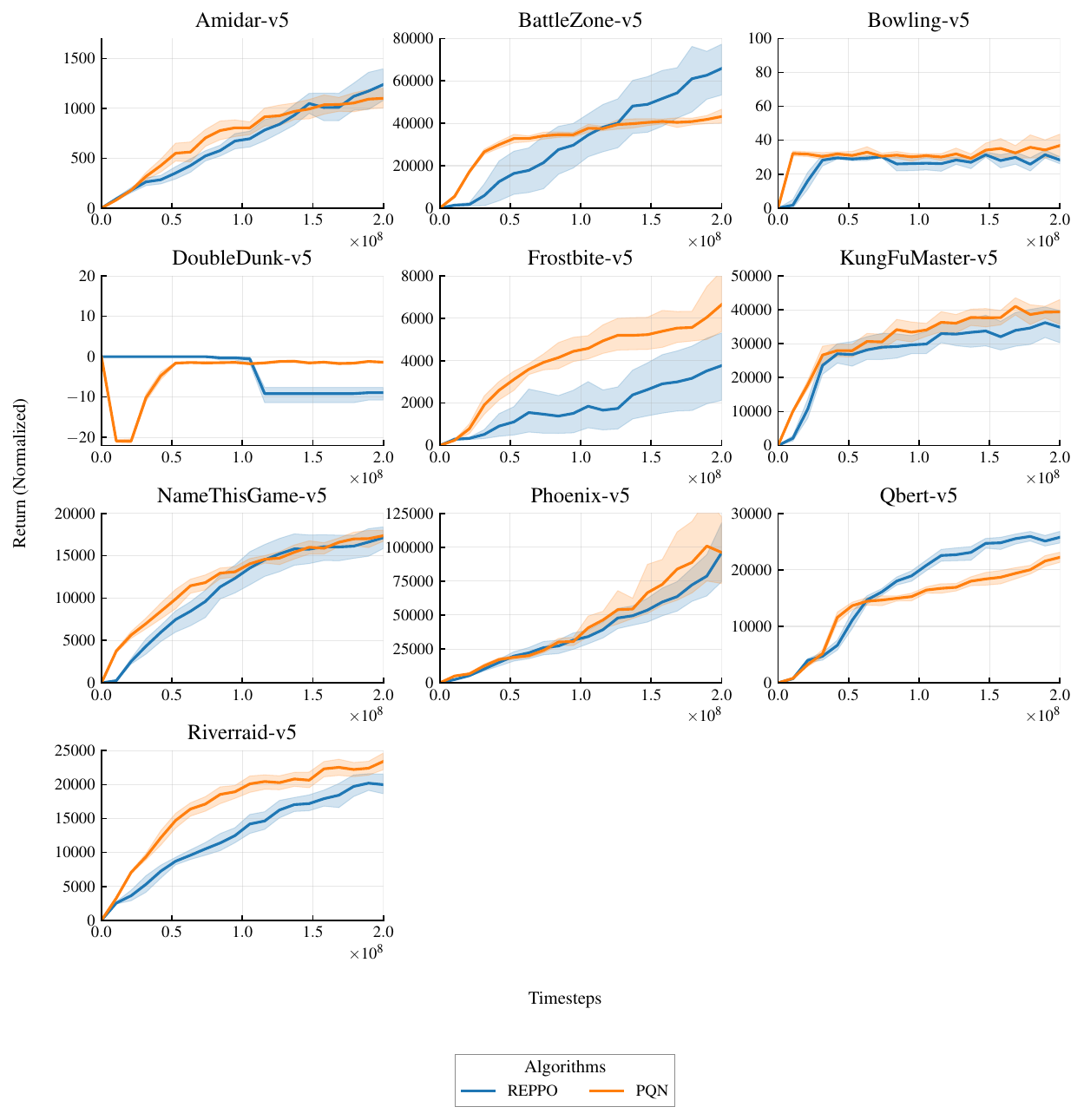}
    \caption{Per-environment results on the Atari-10 suite}
    \label{fig:atari_all}
\end{figure}

We find that our algorithm is able to perform roughly on-par with PQN in the Atari-10 suite of games (cf. \autoref{tab:atari_aggregated} and \autoref{fig:atari_all}) with only minor changes to the architecture to adapt to the Atari games benchmark.
Notably, suitable settings for the KL and entropy target remain consistent even for the discrete action setting.
We only find that the value of $\lambda=0.65$ that is also recommended by \cite{gallici_simplifying_2024} is superior to our default value of $0.95$, likely due to the higher variance of the return in the atari games.
While the high variance across Atari games makes drawing a clear conclusion difficult, we find that PQN seems to achieve slightly better performance.
We find that this is most likely due to the fact that the algorithm adds explicit exploration noise, while we rely on the entropy and conservative KL terms to pace policy improvement.

\begin{table}[h!]
\centering
\caption{Aggregated Human-Normalized Atari-10 scores with 95\% confidence intervals.}
\label{tab:atari_aggregated}
\begin{tabular}{lccc}
\hline
\textbf{Algorithm} & \textbf{Mean [CI]} & \textbf{Median [CI]} & \textbf{IQM [CI]} \\
\hline
REPPO & 2.98 [2.64, 3.33] & 1.68 [1.48, 1.82] & 1.64 [1.54, 1.74] \\
PQN   & 3.35 [3.00, 3.76] & 1.58 [1.48, 1.71] & 1.64 [1.58, 1.71] \\
\hline
\end{tabular}
\end{table}

\section{Implementation details and hyperparameters}
\label{app:example}

In the following, we present implementation details on experiments, as well as a hyperparameter overview.

\subsection{Toy example}

\begin{figure}[t]
    \centering
    \begin{minipage}[]{0.31\linewidth}
    \includegraphics[width=\linewidth]{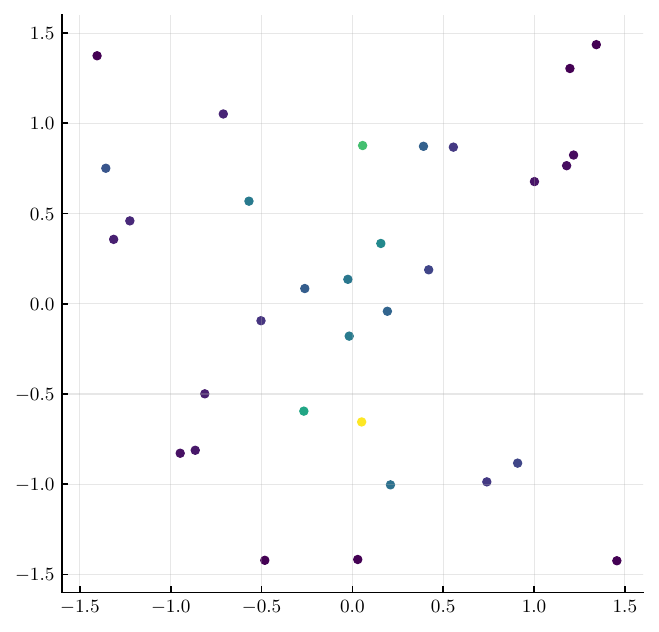}
    \end{minipage}
    \hfill
    \begin{minipage}[]{0.31\linewidth}
    \includegraphics[width=\linewidth]{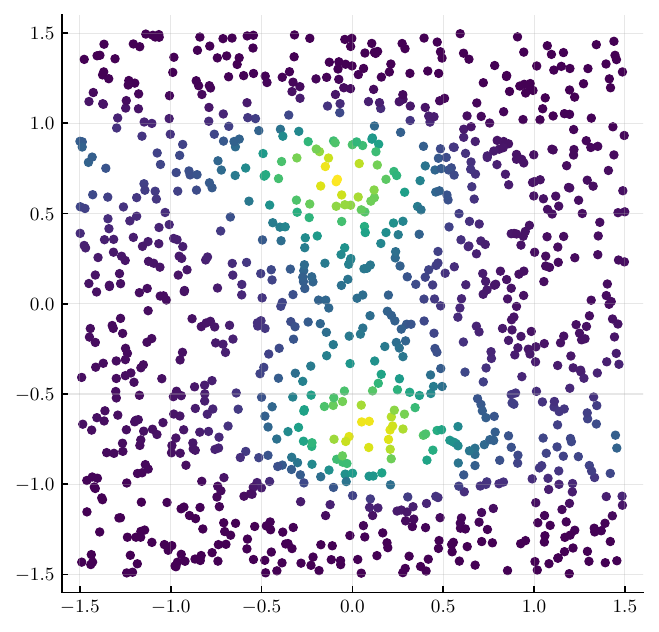}
    \end{minipage}
    \hfill
    \begin{minipage}[]{0.31\linewidth}
    \includegraphics[width=\linewidth]{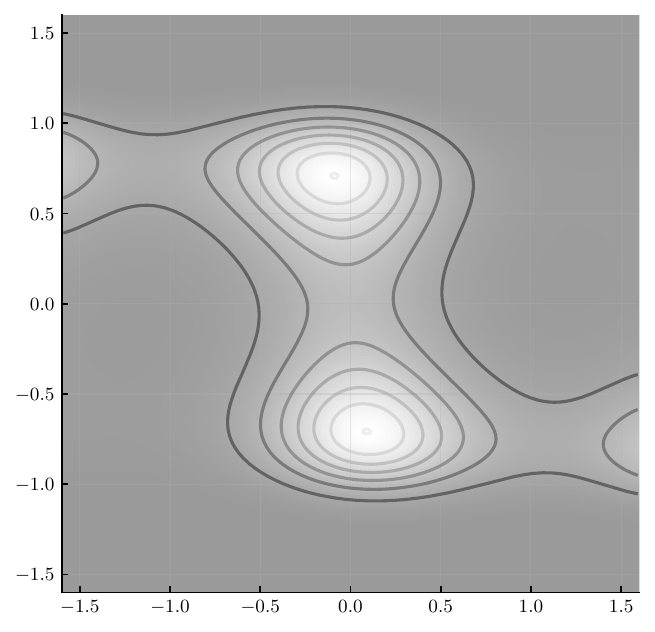}
    \end{minipage}
    \caption{Samples used to train the surrogate function. On the left, we visualize the 32 sample dataset to train the weak surrogate function, in the middle the 1024 datapoints to train the strong, and on the right the full objective function.}
    \label{fig:data_surr_toy}
\end{figure}

To obtain the gradient descent comparison in \autoref{sec:toy} we used the 6-hump camel function, a standard benchmark in optimization.
As our goal was not to show the difficulties of learning with multiple optima, which affect any gradient-based optimization procedure, but rather smoothness of convergence, we initialized all runs close to the global minimum.
The surrogate functions were small three layer, 16 unit MLPs.
To obtain a strong and a weak version, we used differing numbers of samples, visualized in \autoref{fig:data_surr_toy}.
Every algorithm was trained with five samples from the policy at every iteration.
Finally, we tested several learning rates.
We chose a learning rate which allows the ground-truth pathwise gradient to learn reliably.
If a smaller gradient step size is chose, the Monte-Carlo estimator converges more reliably, at the cost of significant additional computation.
We also tested subtracting a running average mean as a control variate from the Monte-Carlo estimate.
While this reduced variance significantly, it was still very easy to destabilize the algorithm by choosing a larger step size or less data samples.

In total, our experiments further highlight a well known fact in gradient-based optimization: while a MC-based gradient algorithm can be tuned for strong performance, it is often extremely dependent on finding a very good set of hyperparameters.
In contrast, pathwise estimators seem to work much more reliably across a wider range of hyperparameters, which corroborates our insights on \algoname hyperparameters robustly transfering across environemnts and benchmark suites.

\subsection{HL-Gauss Equations}
\label{app:gauss}

Given a regression target $y$ and a function approximation $f(x)$, HL-Gauss transforms the regression problem into a cross-entropy minimization. 
The regression target is reparameterized into a histogram approximation $\mathrm{hist}$ of $\mathcal{N}(y,\sigma)$, with a fixed $\sigma$ chosen heuristically. The number of histogram bins $h$ and minimum and maximum values are hyperparameters.
Let $\mathrm{hist}(y)_i$ be the probability value of the histogram at the $i$-th bucket.
The function approximation has an $h$-dimensional output vector of logits.
Then the loss function is
$$\mathrm{HL}(f(x),y) = \sum_{i=1}^h \mathrm{hist}(y)_i \cdot \log \frac{\exp f(x)_i}{\sum_{j=1}^h \exp f(x)_j} \enspace .$$

The continuous prediction can be recovered by evaluating $$\hat{y} = \mathbb{E}[\mathrm{hist}(f(x))] = \langle\mathrm{hist}(f(x)),\mathrm{vec}(\min, \max, h)\rangle,$$ where $\mathrm{vec}(\min, \max, h)$ is a vector with the center values of each bin ranging from $\min$ to $\max$.

\subsection{Auxiliary Task Setup}
\label{app:aux_task}
A simple yet impactful auxiliary task is latent self prediction \citep{schwarzer_data-efficient_2021,voelcker_when_2024,fujimoto_towards_2024}.
In its simplest form, latent self-prediction is computed by separating the critic into an encoder $\phi: \mathcal{X} \times \mathcal{A} \rightarrow \mathcal{Z}$ and a prediction head $f_c: \mathcal{Z} \rightarrow \mathbb{R}$.
The full critic can then be computed as $Q(x,a) = f_c(\phi(x,a))$.
A self-predictive auxiliary loss adds a forward predictive model $f_p: \mathcal{Z} \rightarrow \mathcal{Z}$ and trains the encoder and forward model jointly to minimize 
\begin{align}
\mathcal{L}_\mathrm{aux}(x_t,a_t,x_{t+1},a_{t+1}) = \left|f_p(\phi(x_t,a_t)) - \phi(x_{t+1},a_{t+1})\right|^2.
\end{align}

As our whole training is on-policy, we do not separate our encoder into a state-dependent and action dependent part as many prior off-policy works have done.
Instead we compute the targets on-policy with the behavioral policy and minimize the auxiliary loss jointly with the critic loss.

Overall, the impact of the auxiliary task is the most varied across different environments.
In some, it is crucial for learning, while having a detrimental effect in others.
We conjecture that the additional learning objective helps retain information in the critic if the reward signal is not informative.
In cases where the reward signal is sufficient and the policy gradient direction is easy to estimate, additional training objectives might hurt performance.
We encourage practitioners to investigate whether their specific application domain and task benefits from the auxiliary loss.

\subsection{REPPO Main Experiments}
\label{app:impl}

 \begin{table}[t]
     \centering
     \begin{minipage}[]{0.49\textwidth}
     \begin{tabular}{cc}
  Environment  & \\\hline
  total time steps & $50,000,000$ \\
  n envs & 1024 \\
  n steps & 128 \\
  $\mathrm{KL}_\mathrm{tar}$ & 0.1 \\\hline
  Optimization  & \\\hline
  n epochs &  8\\
  n mini batches & 64\\
  batch size & $\frac{\text{n envs}\, \times\,\text{n steps}}{\text{n mini batches}} = 2048$\\
  lr & $3e-4$ \\
  maximum grad norm & 0.5 \\\hline
  Problem Discount  & \\\hline
  $\gamma$ & $1 - \frac{10}{\max \text{env steps}}$\\
  $\lambda$ & 0.95 \\
  \end{tabular}
     \end{minipage}
\begin{minipage}[]{0.49\textwidth}
  \begin{tabular}{cc}
  Critic Architecture   &\\\hline
  critic hidden dim & 512 \\
  vmin & $\frac{1}{1 - \gamma}\min r $ \\
  vmax & $\frac{1}{1 - \gamma}\max r $ \\
  num HL-Gauss bins & 151 \\
  num critic encoder layers & 2 \\
  num critic head layers & 2 \\
  num critic pred layers & 2 \\\hline
  Actor Architecture  &\\\hline
  actor hidden dim & 512\\
  num actor layers & 3 \\
    \hline
  RL Loss  & \\\hline
  $\beta$ start & 0.01 \\
  $\varepsilon_\text{KL}$ & 0.1 \\
  $\alpha$ start & 0.01 \\
  $\varepsilon_\mathcal{H}$ & $0.5 \times \text{dim} \mathcal{A}$ \\
  aux loss mult & 1.0
     \end{tabular}
\end{minipage}
     \caption{Default \algoname hyperparameters}
     \label{tab:hyper}
 \end{table}

In addition to the details laid out in the main paper, we briefly introduce the architecture and additional design decisions, as well as default hyperparameter settings.

The architecture for both critic encoder and heads, as well as the actor, consists of several normalized linear layer blocks.
As the activation function, we use silu/swift.
As the optimizer, we use Adam.
We experimented with weight decay and learning rate schedules, but found them to be harmful to performance.
Hyperparameters are summarized in \autoref{tab:hyper}.
We tune the discount factor $\gamma$ and the minimum and maximum values for the HL-Gauss representation automatically for each environment, similar to previous work \citep{hansen_td-mpc2_2024}.
This makes the hyperparameters, together with the algorithm description, and the source code, a \emph{complete algorithm specification} in the sense of \cite{jordan2020evaluating}, as we only vary hyperparameters across environments following simple equations on clear, domain sepcific hyperparameters such as the size of the action space and the length of the experiment.

For all environments, we use observation normalization statistics computed as a simple running average of mean and standard deviation.
We found this to be important for performance, similar as in other on policy algorithms.
Since we do not hold data in a replay buffer, we do not need to account for environment normalization in a specialized manner, and can simply use an environment wrapper.

For more exact details on the architecture we refer to interested readers to the codebase.

\section{Additional Results}
\label{app:experiments}

In the following, we provide additional results and further clarification on existing experiments in \autoref{sec:experimental}.

\subsection{Design Ablations}
\label{sec:ablations}

We run ablation experiments investigating the impact of the design components used in \algoname. In these experiments, we remove the cross-entropy loss via HL-Gauss, layer normalization, the auxiliary self-predictive loss, or the KL regularization of the policy updates. To understand the importance of each component for on-policy learning we conduct these ablations for two scales of batch sizes - the default $131,072$ on-policy transitions, as well as the smaller batch size of $32,768$. 

\begin{figure}
\begin{subfigure}{\linewidth}
    \centering
    \raggedleft
    \includegraphics[width=0.91\linewidth]{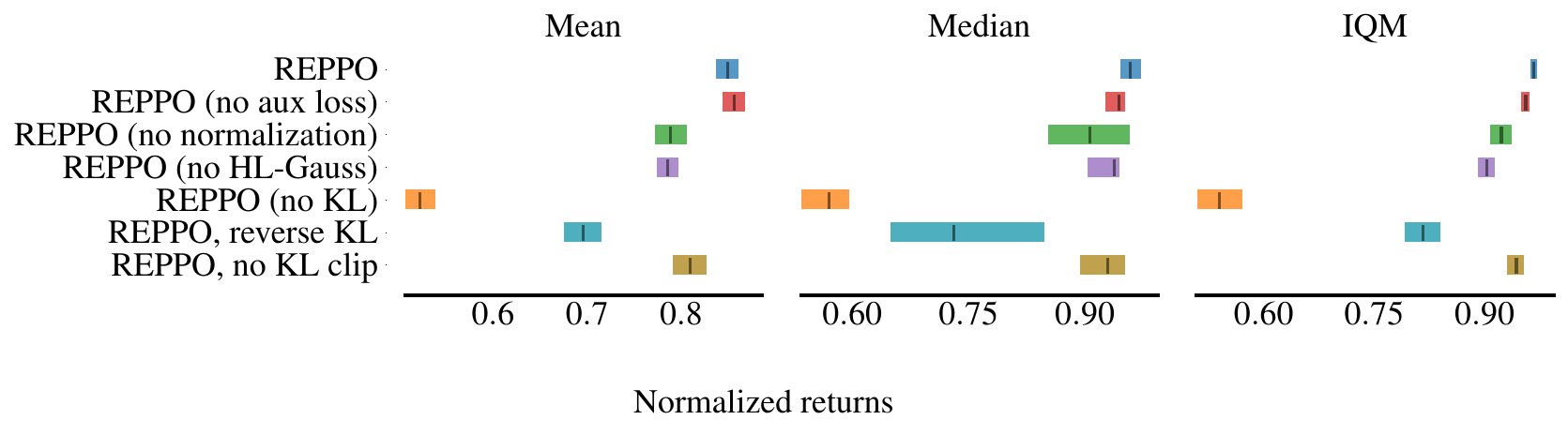}
    \caption{Large dataset size ablation ($128\times1024$).}
\end{subfigure}
\begin{subfigure}{\linewidth}
    \centering
    \includegraphics[width=\linewidth]{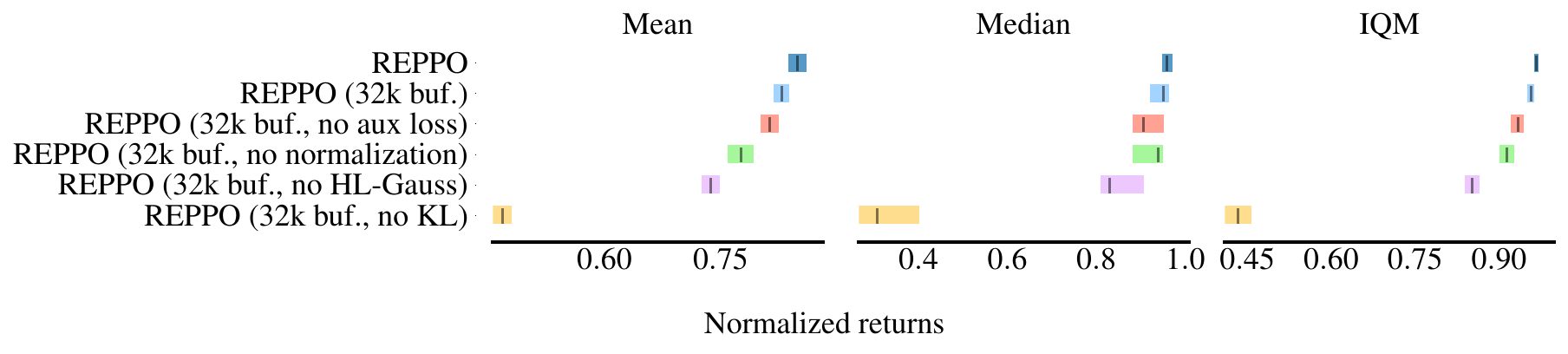}
    \caption{Small dataset size ablation ($32\times1024$).}
\end{subfigure}
    \caption{Ablation on components and data size on the DMC benchmark. Both values are significantly smaller than the replay buffer sizes used in standard off-policy RL algorithms like SAC and FastTD3. The HL-Gauss loss and KL regularization provide a clear benefit at both data scales. The normalization and auxiliary loss become more important when less data is available, highlighting that some stability problems can also be overcome with scaling data.}
    \label{fig:ablation_dmc}
\end{figure}

As shown in \autoref{fig:ablation_dmc}, our results indicate that both the KL regularization of the policy updates and the categorical Q-learning via HL-Gauss are necessary to achieve strong performance independent of the size of the on-policy data used to update our model. 
We find that the KL divergence is the only component that, when removed, leads to a decrease in performance below the levels of PPO, which clarifies the central importance of relative entropy regularization for \algoname.
Removing normalization has minor negative effects on performance which become worse at smaller buffer sizes. 
This is consistent with the literature on layer normalization in RL. 
Similarly, the auxiliary self-predictive loss has a more clearly negative impact on performance when the batch size becomes smaller. 
We note that auxiliary loss has an inconsistent impact on the training generally, where it is strongly beneficial in some environments, but harmful in others.

\subsection{Memory demands and data scaling}

\begin{figure}
\begin{subfigure}{\linewidth}
    \centering
    \begin{tabular}{c|c|c|c|c|c}
         & Num envs & Num steps & Num minibatches & Epochs & Updates per batch \\\hline
        130k buffer & 1024 & 128 & 64 & 8 & 512 \\
        65k buffer & 1024 & 32 & 16 & 8 &  128 \\
        32k buffer & 1024 & 8 & 4 & 8 &  32 \\
        16k buffer & 256 & 8 & 1 & 8 &  8\\
    \end{tabular}
    \caption{Dataset configurations for the data scaling experiment.}
    \label{table:data_comparsion}
\end{subfigure}
\begin{subfigure}{\linewidth}
    \raggedleft
    \includegraphics[width=\linewidth]{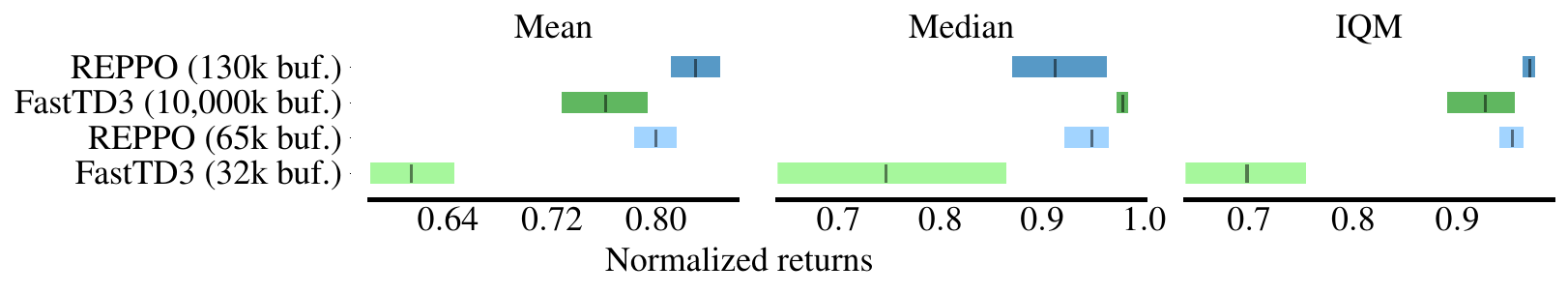}
    \caption{Comparison of aggregate performance between \algoname and FastTD3. \algoname is competitive with the large buffer FastTD3 version and outperforms FastTD3 when memory is limited.}
    \label{fig:fasttd3}
\end{subfigure}

\begin{subfigure}{\linewidth}
    \centering
    \includegraphics[width=\linewidth]{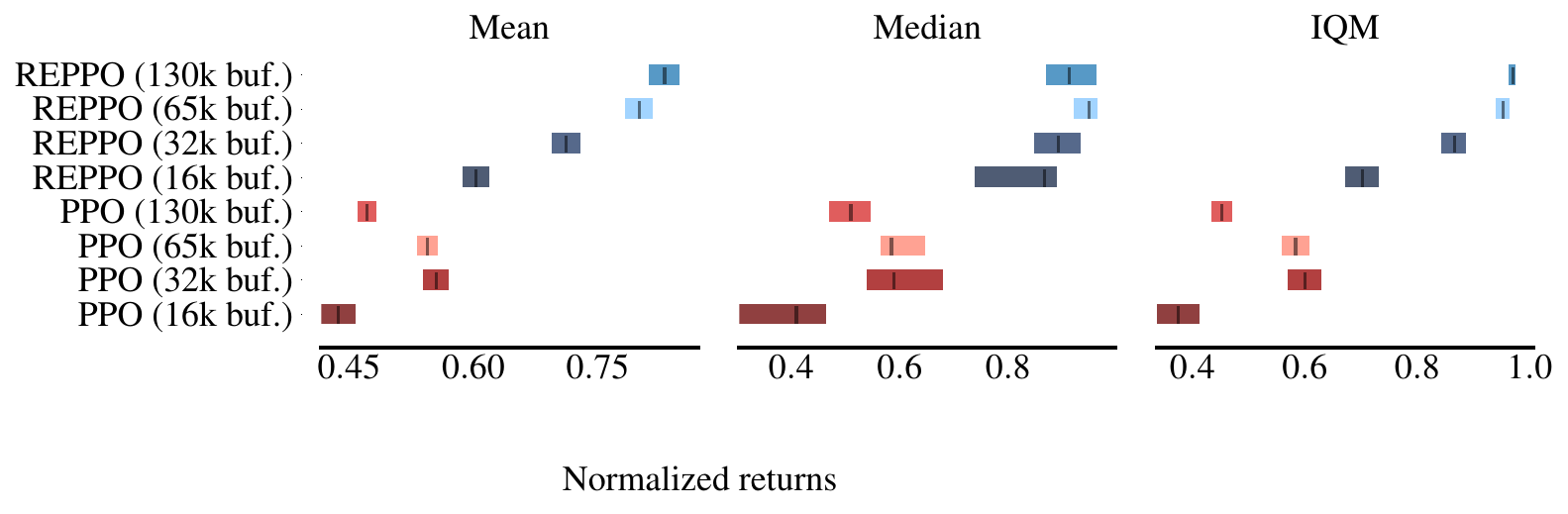}
    \caption{Aggregated performance of REPPO and PPO under different batch dataset sizes. The mean performance of REPPO drops monotonically with decreasing batch size, while PPO shows its highest performance with a medium and small dataset size.}
    \label{fig:data_comparison}
\end{subfigure}
\caption{Experiment to compare the impact of batch datset size on different on-policy algorithms.}
\end{figure}

Recent advances in off-policy algorithms have shown great performance when large buffer sizes are available~\citep{seo2025fasttd3}.
When dealing with complex observations such as images, on-policy algorithms which do not require storing past data have a large advantage.
In terms of data storage requirements, our algorithm is comparable with PPO, yet it remains to answer how well \algoname compares to algorithms that are allowed to store a large amount of data. For this, we compare against the recent FastTD3~\citep{seo2025fasttd3} which also uses GPU-parallelized environments but operates off-policy. We compare \algoname against the original FastTD3 and we also re-run FastTD3 with access to a significantly smaller buffer equivalent to the \algoname buffer. We report the results in \autoref{fig:fasttd3}.

The results demonstrate that \algoname is on par or better in terms of performance on mean and IQM with the FastTD3 approach. This is despite the fact that \algoname uses a buffer that is two to three orders of magnitude smaller. When decreasing the buffer size of FastTD3, the algorithm's performance drops by a large margin while \algoname is barely affected by a smaller buffer. We find that FastTD3 with a smaller buffer can retain performance on lower dimensional, easier tasks but suffers on harder tasks that may be of greater interest in practice. In summary, \algoname is competetive with recent advances in off-policy learning with significantly lower memory and storage requirements.

These results raise the question whether positive scaling with replay buffer size is a general feature of on-policy algorithms.
In our default configuration, REPPO uses long rollouts and a high number of parallel environments, as well as a large number of policy and value function update steps.
PPO on the other hand is often implemented with smaller dataset sizes.
We therefore set up REPPO and PPO training runs across 4 datasets, varying the rollout length.
To keep the total number of gradient steps and the minibatch size the same, we reduced the number of minibatches proportionally to the batch size.
The settings are summarized in \autoref{table:data_comparsion}.
Note that in the large settings, the data becomes more off-policy.
Both PPO and REPPO have explicit ways to deal with this, clipping and the KL minimization term respectively, but the clipping term in PPO is only a heuristic to prevent large importance sampling ratios.

Comparing the performance of both approaches (see \autoref{fig:data_comparison}), we observe a clear pattern.
The mean performance of REPPO drops steeply with decreasing dataset size.
PPO on the other hand does best in the medium and small dataset regimes.
This highlights the different mechanisms on which both algorithms operate.
Larger datasets allow the trained Q function to generalize better, similar to the insight presented in \autoref{fig:pg_comparison}.
On the other hand, for PPO the dataset size needs to be large enough to allow for stable gradient estimation, but not so large that too many gradient update steps are necessary.
This is because clipping can prevent further learning, and many update steps can exacerbate varaince issues with importance sampling.

Note that at some point, REPPO will likely also stop improving with larger datasets and more gradient update steps. 
We see that the performance differences between the medium and the large dataset at not as strong as with smaller datasets.
REPPO cannot continue to learn on fixed data forever, by design, as the KL divergence between two consecutive policies is constrained.
However, we can hypothesize based on the empirical evidence that REPPO is able to scale more gracefully with large amounts of data.

Overall, we can conclude that REPPO scales favorably with larger data buffers, similar to SAC or FastTD3, but is not strongly reliant on it.

\subsection{IsaacLab results}

\begin{figure}
    \centering
    \includegraphics[width=\linewidth]{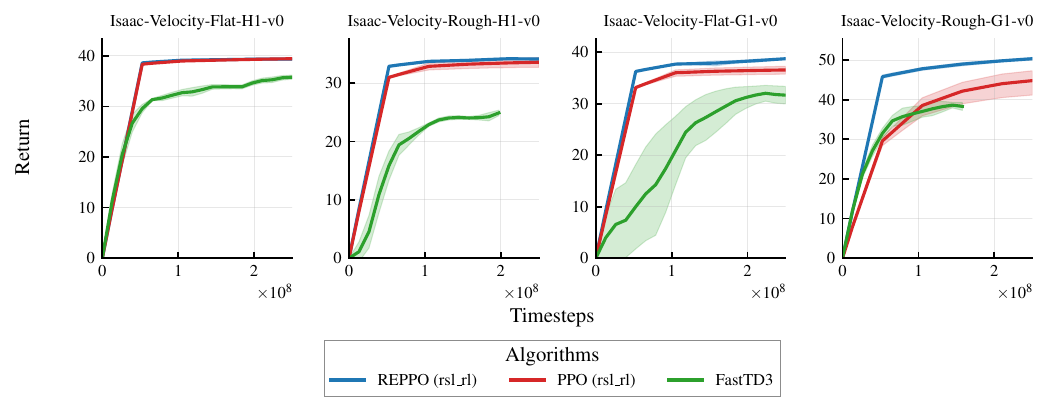}
    \caption{Result overview of REPPO on the IsaacLab locomotion tasks}
    \label{fig:isaac}
\end{figure}

To provide additional value to the robotics community, who often favor specific implementations and APIs for reinforcement learning algorithms, we provide a reimplementation of REPPO in RSL-RL \cite{rslrl}.
We ran this implementation of REPPO and compare the results on the humanoid locomotion environments in the Isaaclab suite \cite{mittal2025isaaclab}.
Results are presented in \autoref{fig:isaac}.
The FastTD3 results provided by \citet{seo2025fasttd3} are not all ran for the full 250 million timesteps.

For IsaacLab, we found that the KL constraint needs to be set more tightly to $0.01$, which is the same value used for PPO.

\subsection{Per Environment Sample Efficiency Curves}
\begin{figure}[b]
    \centering
    \includegraphics[width=\linewidth]{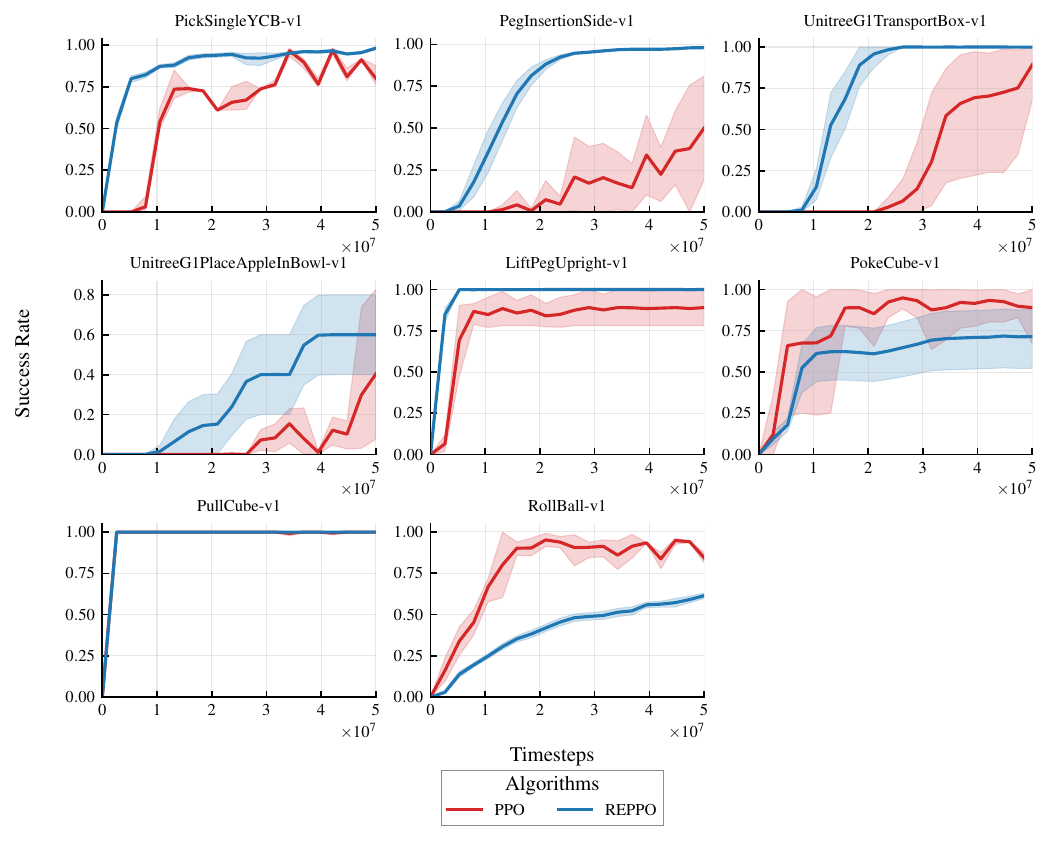}
    \caption{Per-environment results on the ManiSkill suite}
    \label{fig:maniskill_all}
\end{figure}

\begin{figure}[t]
    \centering
    \includegraphics[width=\linewidth]{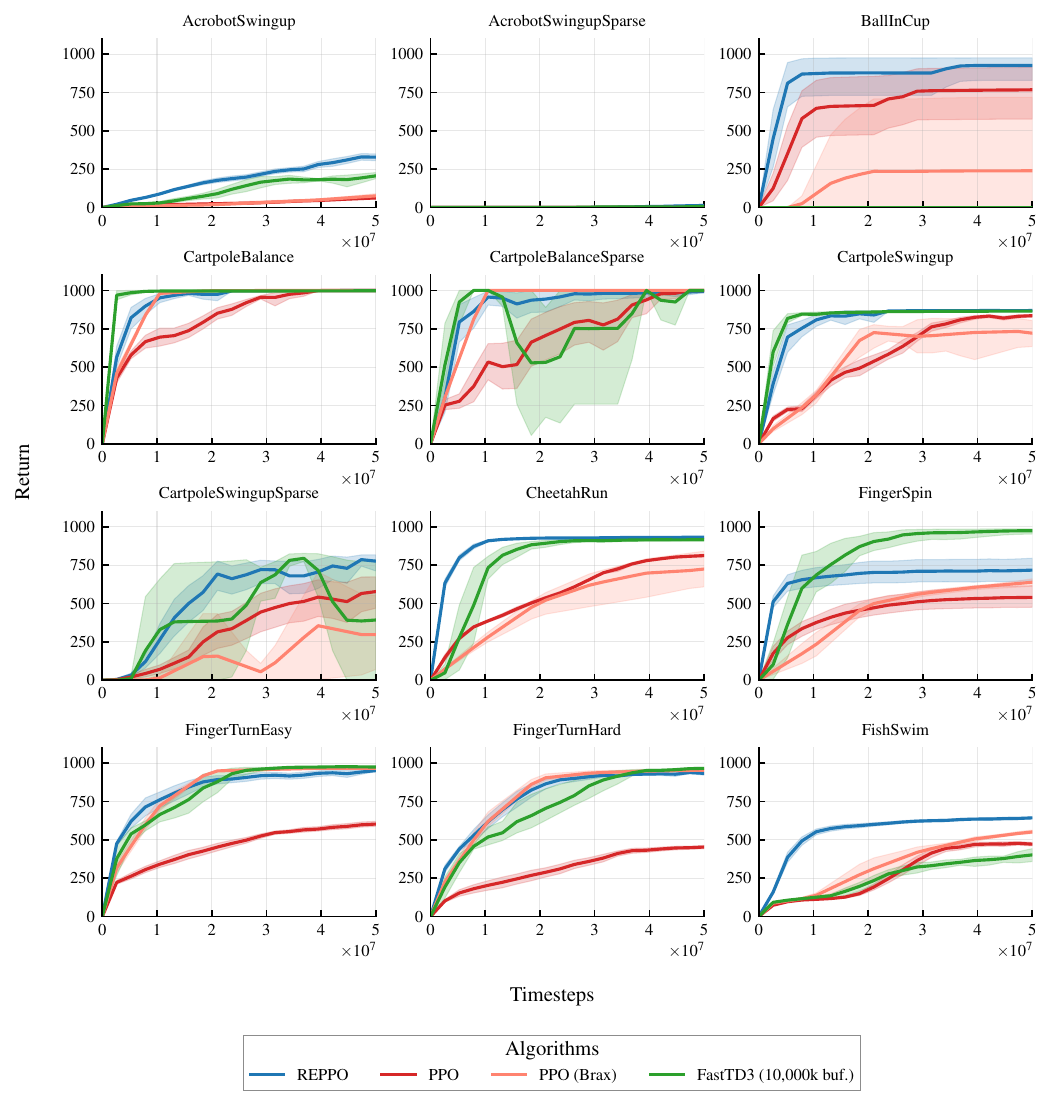}
    \caption{Per-environment results on the \lstinline{mujoco_playground} DMC suite}
    \label{fig:mjx1_all}
\end{figure}

\begin{figure}[t]
    \centering
    \includegraphics[width=\linewidth]{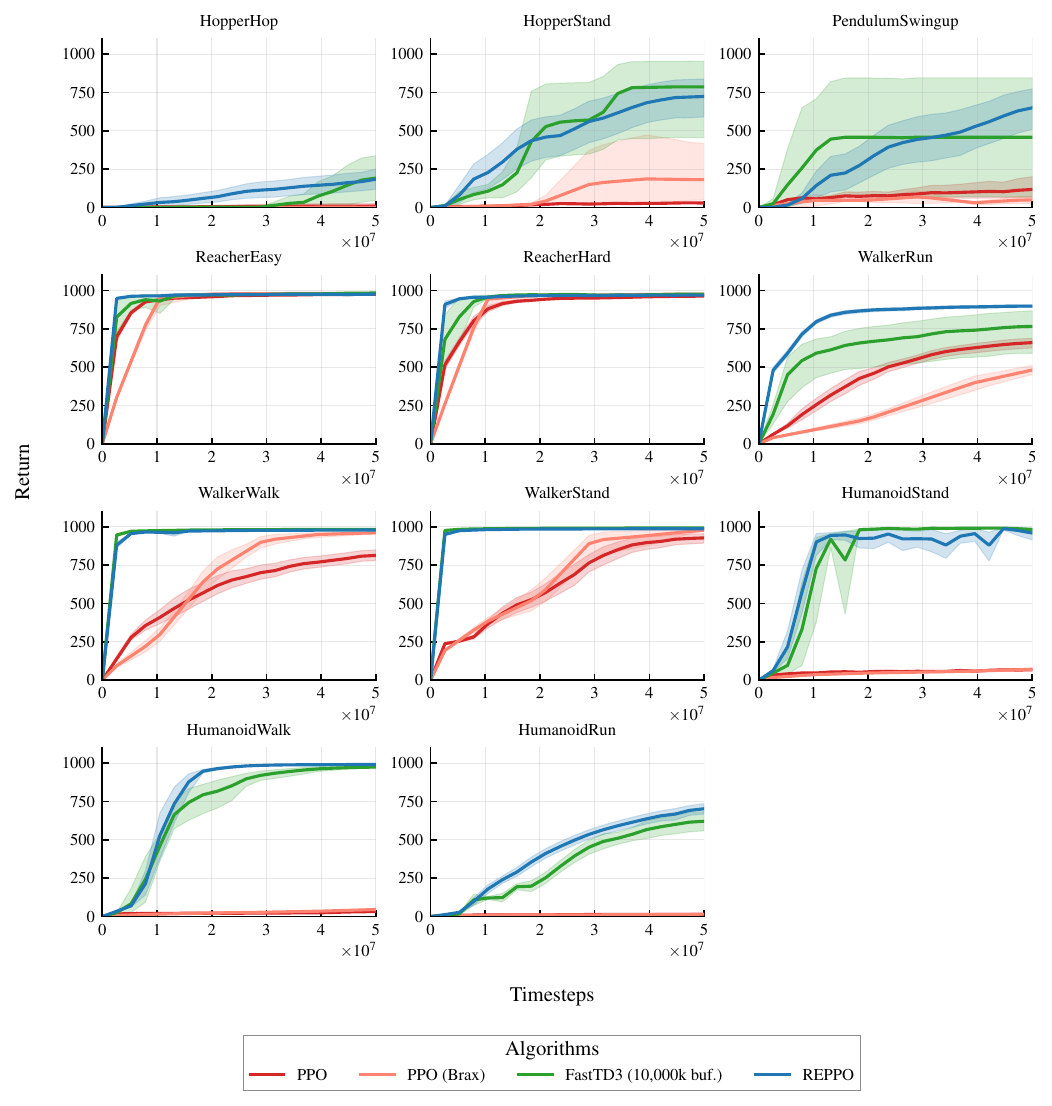}
    \caption{Per-environment results on the \lstinline{mujoco_playground} DMC suite}
    \label{fig:mjx2_all}
\end{figure}

Finally, we provide sample efficiency curves per environment in \autoref{fig:maniskill_all}, \autoref{fig:mjx1_all}, and \autoref{fig:mjx2_all}.

\FloatBarrier
\newpage
\section{Pseudocode}

\begin{algorithm}[h]
\caption{Pseudocode for \fullalgoname}
\label{alg:op-sac}
\KwIn{Environment $\mathcal{E}$, actor network $\pi_\theta$, critic network $Q_\phi$, hyperparameters}
\KwOut{Trained policy $\pi_\theta$}

\SetAlgoLined
\DontPrintSemicolon
\tcp{Initialize networks}
Actor $\pi_\theta$, behavior policy $\pi_{\theta'}$ with $\theta' = \theta$, critic $Q_\phi$ with encoder $f_\phi$, entropy and KL temperature  $\alpha$ and $\beta$  \;

\For{iteration $= 1$ to $N_{iterations}$}{
    \tcp{Step 1: Collect rollout with behavior policy}
    \For{step $= 1$ to $N_{steps}$}{
        \tcp{Apply exploration noise scaling}
        Sample action $a_t \sim \pi_{\theta'}(\cdot|x_t)$ \;
        Execute $a_t$ in environment, observe $(x_{t+1}, r_t, d_t)$ \;
        Compute approximate $V_{t+1} \leftarrow Q_\phi(x_{t+1}, a_{t+1})$ with $a_{t+1} \sim \pi_{\theta'}(\cdot|x_{t+1})$\;
        Compute $\psi_t \leftarrow f_\phi(x_{t+1}, a_{t+1})$
        
        \tcp{Maximum entropy augmented reward, see \autoref{sec:sac}}
        $\tilde{r}_t \leftarrow r_t - \alpha \log \pi_\theta(a_{t+1}|x_{t+1})$ \;
        
        Store transition $(x_t, a_t, \tilde{r}_t, x_{t+1}, d_t, V_{t+1}, \psi_t)$ \;
    }
    
    \tcp{Step 2: Compute TD-$\lambda$ targets, see \autoref{sec:sac}}
    \For{$t = T-1$ down to $0$}{
        $G_t^\lambda \leftarrow \tilde{r}_t + \gamma [(1-d_t)(\lambda G_{t+1}^\lambda + (1-\lambda)V_{t+1})]$ \;
    }
    
    \tcp{Step 3: Update networks for multiple epochs}
    \For{epoch $= 1$ to $N_{epochs}$}{
        Shuffle data and create mini-batches \;
        \For{each mini-batch $b = \{(x,a,G^\lambda,\psi)_i\}_{i=1}^B$}{
            \tcp{Categorical critic update, see \autoref{sec:gauss}}
            $L_Q \leftarrow \frac{1}{B}\sum\text{CrossEntropy}(Q_\phi(x_i,a_i), \text{Cat}(G^\lambda_i))$ \;
            \tcp{Auxiliary task, see \autoref{sec:aux}}
            $L_{aux} \leftarrow \frac{1}{B}\sum||f_\phi(x_i,a_i) - \psi_i||^2]$\;
            Update critic: $\phi \leftarrow \phi - \alpha_Q \nabla_\phi (L_Q + \beta L_{aux})$ \;
            
            \tcp{Actor update with entropy and KL regularization, see \autoref{sec:sac} and \autoref{sec:kl}}
            Sample action $a'_i \sim \pi_\theta(\cdot|x_i)$ \;
            Sample k actions $\bar{a}_i \sim \pi_{\theta'}(\cdot|x_i)$ \;
            Compute KL divergence: $D_\mathrm{KL}(x_i) \leftarrow \sum_{j=1}^k \log \frac{\pi_{\theta'}(\bar{a}_j|x_i)}{\pi_{\theta}(\bar{a}_j|x_i)}$ \;
            
            Policy loss: $L_\pi \leftarrow \frac{1}{B}\sum Q_\phi(x_i,a'_i) - e^\alpha \log \pi_\theta(a'_i|x_i) - e^\beta D_\mathrm{KL}(x_i)$ \;
            (Alternatively, compute clipped objective) \;
            \;
            Update actor: $\theta \leftarrow \theta + \eta_\pi \nabla_\theta L_\pi$ \;
            Entropy $\alpha$ update: $\alpha \leftarrow \alpha - \eta_\alpha \nabla_\alpha e^\alpha(\frac{1}{B}\sum \mathcal{H}[\pi_\theta(x_i)] - \varepsilon_\mathcal{H})$ \;
            KL $\beta$ update: $\beta  \leftarrow \beta - \eta_\beta \nabla_\beta e^\beta (\frac{1}{B}\sum D_{KL}(x_i)] - \varepsilon_\mathrm{KL})$ \;
        }
    }
    
    \tcp{Behavior Policy Update}
    $\theta' \leftarrow \theta$
}
\Return{Trained policy $\pi_\theta$}
\end{algorithm}

\newpage

\end{document}